\documentclass[preprint,12pt, authoryear]{elsarticle}

\usepackage{amssymb}
\usepackage{amsmath}
\usepackage{overpic}
\usepackage{caption}
\usepackage{subcaption}
\usepackage{multirow}
\usepackage{float}
\usepackage{xcolor}
\usepackage{caption}
\usepackage{tabularx}
\usepackage{graphicx} 

\journal{Biomedical Signal Processing and Control}

\begin{document}

\begin{frontmatter}



\title{Dual-Scale Volume Priors with Wasserstein-Based Consistency for Semi-Supervised Medical Image Segmentation}
 \author[label1,label2,label3]{Junying Meng}
\address[label1]{Complex Systems Research Center, Shanxi University, Taiyuan Shanxi, 030006, China}
 \address[label2]{Shanxi Key Laboratory for Mathematical Technology in Complex Systems, Shanxi University,  Taiyuan Shanxi, 030006, China}
 \address[label3]{ Key Laboratory of Complex Systems and Data Science of Ministry of Education, Shanxi University, Taiyuan Shanxi, 030006, China}

 \author[label4]{Gangxuan Zhou}
  \author[label4]{Jun Liu\corref{cor1}}
\address[label4]{Laboratory of Mathematics and Complex Systems (Ministry of Education of China), School of Mathematical Sciences, Beijing Normal University,  Beijing, 100875, China}
\author[label5]{Weihong Guo}
 \address[label5]{Department of Mathematics, Applied Mathematics and Statistics, Case Western Reserve University, Cleveland, 44106, OH, USA}

\cortext[cor1]{Corresponding author.\\ 
E-mail address: jymeng@sxu.edu.cn (J. Meng), 202431130056@mail.bnu.edu.cn (G. Zhou), jliu@bnu.edu.cn (J. Liu), wxg49@case.edu (W. Guo).}

\begin{abstract}

Despite significant progress in semi-supervised medical image segmentation, most existing segmentation networks overlook effective methodological guidance for feature extraction and important prior information from datasets.
In this paper, we develop a semi-supervised medical image segmentation framework that effectively integrates spatial regularization methods and volume priors. 
Specifically, our approach integrates a strong explicit volume prior at the image scale and Threshold Dynamics spatial regularization, both derived from variational models, into the backbone segmentation network.
The target region volumes for each unlabeled image are estimated by a regression network, which effectively regularizes the backbone segmentation network through an image-scale Wasserstein distance constraint, ensuring that the class ratios in the segmentation results for each unlabeled image match those predicted by the regression network.
Additionally,  we design a dataset-scale Wasserstein distance loss function based on a weak implicit volume prior, which enforces that the volume distribution predicted for the unlabeled dataset is similar to that of labeled dataset. Experimental results on the 2017 ACDC dataset, PROMISE12 dataset, and thigh muscle MR image dataset show the superiority of the proposed method.

\end{abstract}




\end{frontmatter}



\section{Introduction}
\label{sec:introduction}
In image processing, segmentation is an important task that partitions the pixels in an image to form several semantically meaningful regions, where each region consists of pixels with similar features or attributes.
Image segmentation methods are generally categorized into model-based and learning-based approaches. 
Traditional variational segmentation models generally consist of two components: data fidelity and regularization. The data fidelity term quantifies the similarity within one region and the difference between different regions, facilitating the delineation of boundaries. The additional regularization term is used to incorporate expectations on the results by assuming that the derived segmentations adhere to a proper function space. Examples of regularization terms include boundary smoothness, shape similarity, and volume consistency. Total variation (TV) regularization,  introduced by \citep{TV}, is widely adopted to improve segmentation accuracy and robustness \citep{Li}. However, despite the development of many fast algorithms \citep{TVa1, TVa3}, TV regularization is nonsmooth and remains challenging to be solved efficiently.
In addition, pre-selected regularity in model-based methods might not work for the images at hand.
Recently, data-driven approaches, especially deep convolutional networks, have demonstrated great success across numerous tasks in computer vision, owing to their powerful feature extraction abilities.
Specifically, for image segmentation tasks, Deep Convolutional Neural Networks (DCNNs) have achieved significant success. This impressive performance typically relies on large-scale annotated datasets.
However, obtaining such precise annotations is often challenging and expensive, especially for medical images, which commonly suffer from low contrast and noise.
As a result, novel deep segmentation frameworks have been proposed to handle scenarios with insufficient labeled samples.

Semi-supervised segmentation leverages limited annotated data along with abundant unlabeled data to enhance training.
There are various strategies for utilizing partially annotated data. Self-training is a common strategy \citep{pslabelwork1, pslabelwork4, pslabelwork3, pslabelwork2}, which starts with an initial model trained on a small labeled dataset. The model then generates pseudo masks for unlabeled images, and these pseudo masks are combined with the ground truth labels to refine the model iteratively.
Although this method leverages unlabeled images, a major drawback is that the segmentation results based on limited annotations may contain noise. 
Directly treating these noisy predictions as pseudo-labels may introduce misleading supervision during training and compromise overall performance \citep{pslabel}.
Consistency learning is one way to enhance the invariance of predicted outputs by applying different perturbations to input images. 
The mean teacher model \citep{meanteacher} is a widely adopted semi-supervised learning approach that enforces data consistency by adding noise to the inputs and updating the teacher model's parameters via exponential moving averages.
UA-MT \citep{utma} builds on this framework by incorporating dropout to filter unreliable outputs from the teacher, thereby enabling more stable and effective training of the student model.
MC-Net \citep{mco} is a model that uses two decoders with different upsampling techniques and incorporates recurrent soft pseudo-labels to create a mutual consistency model.
They further introduced an enhanced version called MC-Net+ \citep{mc}, which incorporates three decoders to effectively utilize the information differences among various methods for cycle consistency learning.
In addition, other techniques include co-training \citep{co4, co5, liux2022b, co2} and adversarial learning \citep{adv2, adv4, adv6, adv3, adv1}. However, these strategies often require training multiple networks or employing multiple objective functions, which may impede convergence.
Distribution prior-based semi-supervised segmentation methods encourage the model to generate prediction distributions that align with the ground truths distributions. For example, an adversarial network was proposed in \citep{hung2019adversarial} to encourage segmentation predictions for unlabeled data to align with the distribution of ground truth labels.
A GAN-based architecture that leverages distribution priors to extract knowledge from large amounts of unlabeled data was proposed in \citep{souly2017semi}. While these approaches align the overall distributions of predictions and ground truths, they often overlook the alignment of specific features, such as volume distributions.


Variational image segmentation models have been shown to significantly improve segmentation performance by incorporating regularization priors \citep{mumford1989optimal, pott}. Volume constraints are an important prior in many practical image segmentation applications.
In scenarios involving poor-quality images affected by noise, blur, or shading, preserving volume is crucial for accurate segmentation.
Additionally, in medical image segmentation, leveraging volume priors obtained from experience or other sources can significantly improve segmentation outcomes.
For instance, the MBO scheme incorporating volume constraints \citep{volume1} has been applied to segmentation in \citep{volume}.
Their numerical experiments demonstrate that enforcing preserving volume can notably improve segmentation accuracy, especially when training data is scarce. Importantly, this advantage persists even with coarse volume bound estimates. 
DCNNs such as U-Net \citep{unet}, Seg-Net \citep{segnet}, and DeepLabv3+ \citep{deeplab} have proven highly effective for image segmentation. 
However, these network architectures are generally limited in their ability to incorporate spatial priors. 
To overcome this limitation, a Soft Threshold Dynamics (STD) segmentation method was proposed \citep{liuj2022a}, which significantly improves DCNNs by incorporating various priors, including spatial regularity, volume constraints, and star-shape priors.

Some semi-supervised segmentation methods have incorporated volume constraints.
For example, a curriculum-style design incorporating a main and auxiliary network was introduced in \citep{curr} for semi-supervised segmentation tasks.
The auxiliary network is designed to predict the volume of the target region. 
Similarly, a curriculum-style strategy for multi-task semi-supervised segmentation was introduced in \citep{curr1}.
However, these methods incorporate volume constraints into semi-supervised segmentation via an auxiliary network, rather than integrating them into the segmentation network architecture.  
Moreover, they only consider volume constraints at the image scale, overlooking volume distribution priors at the dataset scale. As a result, they fail to fully leverage the overall statistical information of the dataset.

In this paper,  we introduce a semi-supervised network architecture for medical image segmentation. Specifically, the backbone segmentation network within this framework incorporates a strong volume prior and Threshold Dynamics (TD) spatial regularization. 
For unlabeled images, a regression network is employed to predict the volumes of target regions. These predictions help regularize the segmentation network by enforcing consistency between the class ratios in its softmax output and those predicted by the regression network for each unlabeled image. 
Specifically, at the image scale, the volume for each image is explicit and can be represented by the class ratio. 
To enforce this consistency, we introduce a loss function calculated from the Wasserstein distance across the two explicit distributions.
Unlike previous methods that utilize volume priors for semi-supervised segmentation, we also consider the volume distribution priors at the dataset scale. Specifically, the volume of each image is treated as a sample from the dataset's volume distribution. This dataset-scale weak prior is implicit. 
Therefore, we employ a Wasserstein distance to enforce consistency of volume distributions from ground truth and predictions on unlabeled data.
Experiments show that the proposed method effectively integrates volume priors and achieves superior performance
 compared to several leading approaches.

Overall, the main contributions of this work are as follows:
\begin{itemize}
\item The proposed semi-supervised segmentation framework innovatively incorporates strong image-scale volume priors, derived from a mathematical variational model, into the backbone segmentation network. This design enhances the mathematical interpretability of the segmentation process and facilitates more reliable analysis of tissues and organs. 

\item 
 To address the limitations of fixed volume estimates from previous methods, we introduce a learnable regression network that predicts the volume of unlabeled images, allowing dynamic adaptation to each individual image. 
Furthermore, a Wasserstein loss is used to align the predicted class ratios from the regression network with the segmentation output for each unlabeled image, effectively regularizing the backbone segmentation network.

\item  We innovatively introduce a weak dataset-scale volume distribution prior by treating the each image volume as a sample from the dataset’s volume distribution, and applying a Wasserstein loss to enforce consistency between the predicted volume distributions for the unlabeled data and those derived from labeled data.

\end{itemize}


The rest of this paper is organized as follows: Section \ref{relatedwork} reviews some related work; Section \ref{proposed} presents proposed network architecture that integrates a strong image-scale volume prior and introduces a loss function based on a weak dataset-scale volume distribution prior. Section \ref{experiment} provides experimental results; Section \ref{conclusion} concludes the paper and offers insights for future work.

\section{The related work}
\label{relatedwork}
\subsection{Potts model}
The Potts model \citep{pott} is a classic variational image segmentation approach. Assume the image domain $\Omega$ is partitioned into $K$ non-overlapping subdomains $\{\Omega_{k}\}_{k=1}^{K}$.
Let $I: \Omega\subset \mathbb{R}^{2}\rightarrow \mathbb{R}^{d}$ represent an image defined on $\Omega$,
where $d=1$ corresponds to grayscale images, and $d=3$ to color images. 
Let $\Omega=\{x_{n}\}_{n=1}^{N}$ be the discrete set representing all pixels in image $I$, where $N$ is the total pixel count, and let $K$ represent the number of segmentation classes. 
A relaxed formulation of the Potts model is given by:
\begin{equation*}\label{equ5}
\mathop{\min}_{\mathbf{h}\in\mathbb{H}}\sum_{k=1}^{K}\sum_{n=1}^{N}c_{k}(x_{n})h_{k}(x_{n})+\lambda \sum_{k=1}^{K}\sum_{n=1}^{N}\| \nabla h_{k}(x_{n})  \|,
\end{equation*}
where $c_{k}(x_{n})$ measures how closely pixel $I (x_{n})$ matches the $k$-th segmentation class.
The second term corresponds to the classical TV regularization and quantifies the overall boundary length of the segmented regions, assuming that $h_{k}$ denotes the indicator function of the $k$-th region $\Omega_{k}$. The parameter $\lambda\geqslant 0$ controls the trade-off between the data fidelity and regularization terms. The simplex set 
\begin{equation*}
    \begin{array}{rl}
    \mathbb{H}=\left\{\mathbf{h}=(h_{1},...,h_{K}):  \sum_{k=1}^{K}h_{k}(x_{n})=1,\right.\\
    \left. h_{k}(x_{n})\in[0,1], \forall x_{n} \in \Omega, k=1,...,K.\right\}
    \end{array}
\end{equation*}
forms the segmentation condition.

\subsection{Volume preserving soft threshold dynamics}

To overcome the lack of volume prior in the segmentation condition $\mathbb{H}$ in Potts model, a volume constraint was introduced \citep{liuj2022a} by modifying $\mathbb{H}$ as 
$$\mathbb{H}_\mathbf{V}=\{\mathbf{h}\in\mathbb{H}, \sum_{n=1}^{N}h_{k}(x_{n})=V_{k}\},$$
where $\mathbf{V} = (V_{1}, V_{2}, ..., V_{K})$ represents the given volumes, with $V_{k}$ denoting the volume of the $k$-th class. Instead of using TV regularization, they incorporated a smooth TD spatial regularization term $\mathcal{R}(\mathbf{h})$ to penalize boundary lengths. This spatial prior contributes to improve segmentation performance by encouraging smoother boundaries and enhancing robustness to noise. Consequently, they formulated the following  Soft Threshold Dynamics (STD) variational segmentation model:
\begin{equation}\label{model}
\begin{aligned}
\mathop{\min}_{\mathbf{h}\in\mathbb{H}_{\mathbf{V}}}\underbrace{\sum_{k=1}^{K}\sum_{n=1}^{N}c_{k}(x_{n})h_{k}(x_{n})+\varepsilon\sum_{k=1}^{K}\sum_{n=1}^{N}h_{k}(x_{n})\ln h_{k}(x_{n}) }_{: =\mathcal{F}(\mathbf{h};\mathbf{c})}\\
+ \underbrace{\lambda\sum_{k=1}^{K}\sum_{n=1}^{N}h_{k}(x_{n})(k*(1-h_{k}))(x_{n})}_{: =\mathcal{R}(\mathbf{h})},
\end{aligned}
\end{equation}
where $\varepsilon\geqslant 0$ is an entropic regularization parameter, and $*$ denotes the convolution operation.

Given an input image $I$, let $I =\mathbf{m}^{0}$. The DCNN $\mathcal{S}_{\theta}$, parameterized by $\theta$ for image segmentation, can be expressed as $\mathbf{m}^{T}=\mathcal{S}_{\theta}(\mathbf{m}^{0})$.
Inspired by the STD variational segmentation model, the $T$ layers of the DCNN can be reformulated as 
\begin{equation*}\label{dncnn1}
\left\{
\begin{aligned}
&\mathbf{o}^{t}=\mathcal{T}_{\theta^{t-1}}(\mathbf{m}^{t-1},\cdots,\mathbf{m}^{0}),\\
&\left\{
\begin{aligned}
\mathbf{m}^{t}&=\mathcal{A}^{t}(\mathbf{o}^{t}),
t=1,2,\cdots,T-1.\\
\mathbf{m}^{T}&=\mathcal{A}^{T}(\mathbf{o}^{T})=\mathop{\arg\min}_{\mathbf{h}\in\mathbb{H}_{\mathbf{V}}}\{\mathcal{F}(\mathbf{h};-\mathbf{o}^{T})+\mathcal{R}(\mathbf{h})\}.
\end{aligned}
\right.
\end{aligned}
\right.
\end{equation*}
Here, $\mathcal{T}_{\theta^{t-1}}$ is an operator that connects the $t$-th layer $\mathbf{o}^{t}$ to the previous layers $\mathbf{m}^{t-1},\cdots,\mathbf{m}^{0}$, $\mathcal{A}^{t},t=1,2,\cdots,T-1$ represent the activation functions, and the final $\mathcal{A}^{T}$
is derived by solving the primal STD variational problem (\ref{model}) using a dual algorithm. Specifically, the equivalent dual problem of (\ref{model}) is given by:
$$\mathbf{u}^*=\mathop{\arg\max}_{\mathbf{u}\in\mathbb{H}}\{\langle \mathbf{u}, \mathbf{V}\rangle + \langle \mathbf{u}^{c,\varepsilon},\mathbf{1} \rangle\},$$
where $\mathbf{u}^{c,\varepsilon}$ denotes the $\varepsilon$-entropically regularized c-concave transform of $\mathbf{u}$, and $\mathbf{u}^{*}$ denotes the optimizer of this dual problem.
Then, the optimal solution $\mathbf{h}^{*}$ to the primal problem  can be derived as:
\begin{equation*}
\mathbf{h}^{*}=softmax\big(g(\mathbf{o}^{T}, \mathbf{u}^{*},\mathbf{V})\big),
\end{equation*}
where $g$ describes the functional relationship between $\mathbf{o}^{T}$, $\mathbf{u}^{*}$ and $\mathbf{V}$.
Then the final softmax classification function 
$\mathbf{m}^{T}=\mathbf{h}^{*}$  can be determined, where each component $m_{k}^{T}(x_{n})$ represents the probability of assigning pixel $x_{n}$ to class $k$. The detailed 
algorithm for finding $\mathbf{h}^*$ and the related unrolling subnetworks can be found in \citep{liuj2022a}.

Based on the STD variational segmentation model, the network $\mathcal{S}_{\theta}$ can be expressed as 
\begin{equation}\label{u}
\mathbf{m}^{T}=\mathcal{S}_{\theta}(I, \mathbf{V}).
\end{equation}
Its final layer is a VP-STD softmax layer, which integrates Volume Preservation (VP) and STD spatial regularization to enhance segmentation accuracy.
During training, the volume $\mathbf{V}$ in (\ref{u}) is directly obtained from the available annotated labels in the training dataset. 
However, during testing, ground truth labels are unavailable. Therefore, the average volume $\mathbf{V}_{emp}$ of the ground truth labels in the training set serves as an approximation for the test image volume. The segmentation process is represented by:
\begin{equation}\label{emp}
\mathbf{m}^{T}=\mathcal{S}_{\theta}(I, \mathbf{V}_{emp}).
\end{equation}
This strategy provides a rough estimate of the volume, and using the same volume prior across different test images in the softmax layer is unreasonable.

\section{The proposed network with dual-scale volume priors}
\label{proposed}
\subsection{The proposed strong image-scale volume prior}

In this section, we incorporate both the TD spatial regularization prior and the strong image-scale volume prior into the semi-supervised segmentation framework. 
To address the limitation of relying on a coarse, fixed estimate such as $\mathbf{V}_{emp}$, we propose a learnable volume prediction mechanism that dynamically infers the expected volume for each input image.
This approach enables the incorporation of adaptive, image-specific volume priors, effectively avoiding rough approximations and providing more accurate volume constraints to guide the segmentation process.

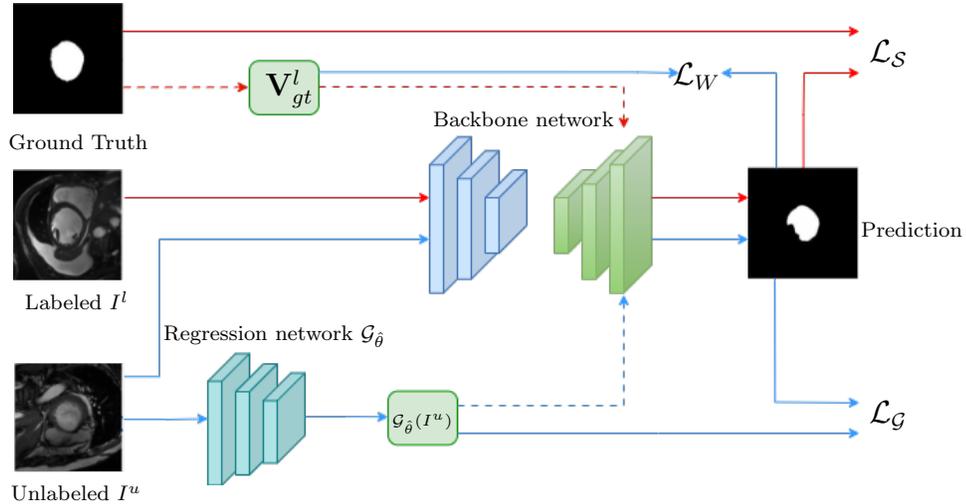
\begin{figure*}[htbp]
\centering         
\begin{overpic}[width=0.83\textwidth]{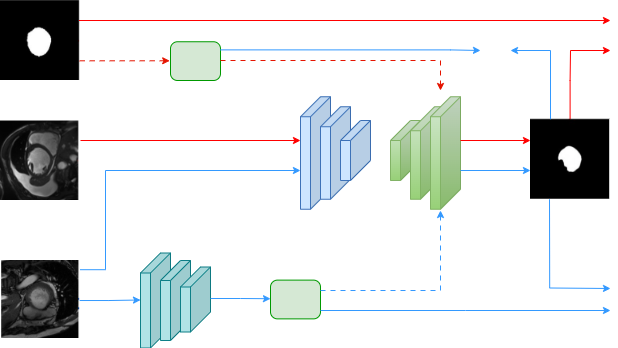}

\put(0.1, 20.5){ \scriptsize{Labeled $I^{l}$}}

\put(-1.2, -1.8){\scriptsize{  Unlabeled $I^{u}$}}

\put(-0.6, 39.3){\scriptsize{Ground Truth}}
\put(99, 29){\scriptsize{Prediction}}
\put(100, 49.5){$\mathcal{L}_{\mathcal{S}}$}
\put(100, 7.3){$\mathcal{L}_{\mathcal{G}}$}
\put(77, 46.8){$\mathcal{L}_{W}$}
\put(44.2, 6.9){\tiny
{$\mathcal{G}_{\hat{\theta}}(I^{u})$}}
\put(49, 42){\scriptsize{Backbone network}}
\put(17.5, 17){\scriptsize{Regression network $\mathcal{G}_{\hat{\theta}}$}}
\put(29.4, 45.5){$\mathbf{V}_{gt}^{l}$}
\end{overpic}

\caption{Illustration of the proposed semi-supervised segmentation network VP-Net. Red and blue arrows indicate the processing paths for labeled and unlabeled data, respectively. $\mathcal{L}_{\mathcal{S}}$, $\mathcal{L}_{\mathcal{G}}$, and $\mathcal{L}_{W}$ denote the supervised loss, the regularization loss, and the weak volume distribution prior loss, respectively.}\label{network}
\end{figure*}

Considering a semi-supervised segmentation scenario with two subsets: $\mathbb{L}=\{(I^{l}(x_{n}),G^{l}(x_{n})) | x_{n}\in\Omega\}_{l=1,...,L}$ 
which consists of a set of images $\{I^{l}\}_{l=1,...,L}$ together with pixel-level ground truth $\{G^{l}\}_{l=1,...,L}$, and $\mathbb{U}=\{I^{u}(x_{n})| x_{n}\in\Omega\}_{u=1,...,U}$, which is a set of unlabeled images.
Our semi-supervised segmentation framework, VP-Net, is illustrated in Fig. \ref{network}. By incorporating the VP-STD softmax layer as the final layer of the backbone segmentation network, VP-Net effectively integrates the volume prior and TD spatial regularization prior from the variational model. Unlike the frameworks in \citep{curr, curr1} using auxiliary networks to impose volume constraints, our approach directly integrates priors derived from a mathematical variational model into the semi-segmentation network architecture, thereby improving both model's performance and interpretability.

For labeled images, the segmentation process is represented by the red line in Fig. \ref{network}.
The ground truth volume of the labeled image $I^{l}$, denoted as $\mathbf{V}_{gt}^{l}$, is fed into the final VP-STD softmax layer. Thus, the supervised segmentation for labeled image $I^{l}$ can be expressed as:
\begin{equation*}\label{ugt}
\mathbf{m}^{T}=\mathcal{S}_{\theta_{1}}( I^{l}, \mathbf{V}_{gt}^{l}),
\end{equation*}
where $\mathcal{S}$ denotes the backbone segmentation network parameterized by $\theta_{1}$.
The loss $\mathcal{L}_{\mathcal{S}}$ for labeled images is formulated using the multi-class cross-entropy as follows:
  \begin{equation*}\label{segloss}
\mathcal{L}_{\mathcal{S}}(\theta_{1})=-\frac{1}{N}\sum_{l=1}^{L}\sum_{n=1}^{N}\sum_{k=1}^{K}G_{k}^{l}(x_{n})\ln \mathcal{S}_{\theta_{1}}(I^{l}(x_{n}), \mathbf{V}_{gt}^{l})_{k},
  \end{equation*}
where $\mathcal{S}_{\theta_{1}}(I^{l}(x_{n}), \mathbf{V}_{gt}^{l})_{k}$ represents the softmax output of the backbone segmentation network $\mathcal{S}$, representing the probability that the pixel at location $x_{n}$ in image $I^{l}$ belongs to class $k$, and $G_{k}^{l}(x_{n})$ denotes the ground truth label for the pixel at location $x_{n}$ with respect to class $k$.

For unlabeled images $\{I^{u}\}_{u=1,...,U}$, where the ground truth labels are unknown, the volume prior cannot be directly determined. 
In this work, we propose a learnable regression network $\mathcal{G}$, parameterized by $\hat{\theta}$, which predicts image-level features, specifically estimating the volume of target regions in unlabeled images.
The predicted volume $\mathcal{G}_{\hat{\theta}}(I^{u})$ for $u=1,...,U$ is subsequently incorporated into the VP-STD softmax, enabling the model to dynamically adapt to each unlabeled sample. Thus, the unsupervised segmentation for unlabeled image $I^{u}$ can be expressed as:
\begin{equation*}
\mathbf{m}^{T}=\mathcal{S}_{\theta_{1}}(I^{u}, \mathcal{G}_{\hat{\theta}}(I^{u})).
\end{equation*}
The corresponding segmentation process is illustrated in Fig. \ref{com}(b).
In contrast to the segmentation (\ref{emp}) in \citep{liuj2022a}, which uses a fixed empirical volume calculated from the training set’s ground truth labels (as shown in Fig. \ref{com}(a)), our approach introduces a learnable volume prediction mechanism that dynamically adapts to each sample. This method avoids using a fixed volume  and provides more flexible and accurate volume predictions.

\setcounter{figure}{1} 
\begin{figure}[htbp]
\centering

\begin{minipage}{0.63\textwidth} 
\centering
\begin{overpic}[width=\textwidth]{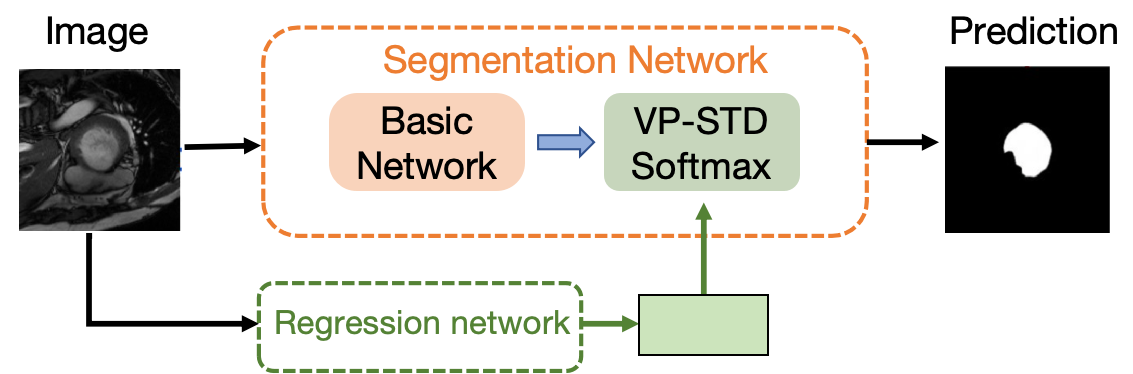}
\put(57, 3.6){ \tiny{$\mathbf{V}_{emp}$}}  
\end{overpic}
\vspace{2pt} 
\subcaption{Fixed volume estimation \citep{liuj2022a}.}\label{network-approx}
\end{minipage}

\begin{minipage}{0.63\textwidth} 
\centering
\begin{overpic}[width=\textwidth]{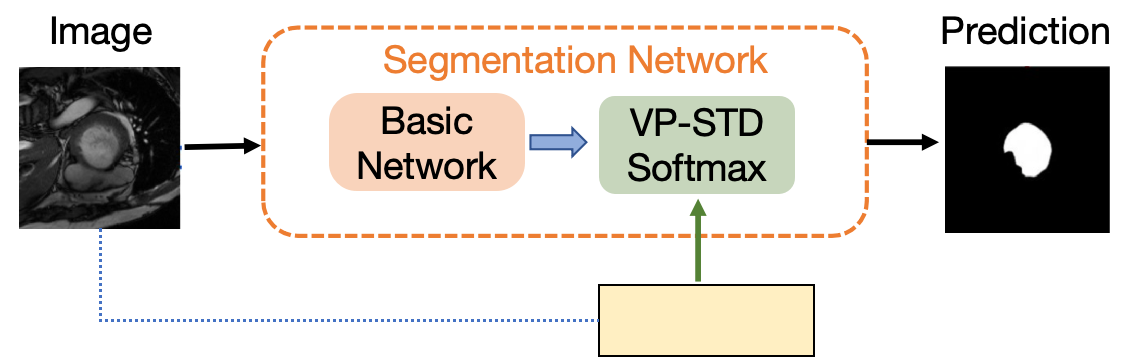}
\put(56, 2.5){ 
\tiny{$\mathcal{G}_{\hat{\theta}}(I^{u})$}}
\end{overpic}
\vspace{2pt} 
\subcaption{Proposed learnable volume prediction.}\label{network-learnable}
\end{minipage}

\vspace{5pt} 
\caption{Segmentation networks incorporating VP-STD softmax for unlabeled images.}\label{com}
\end{figure}


We first train the regression network $\mathcal{G}$ with parameters $\hat{\theta}$ through optimization of the following objective function:
$$ \mathop{\min}_{\hat{\theta}}\sum_{l=1}^{L}\sum_{k=1}^{K}(\mathcal{G}_{\hat{\theta}}(I^{l})_{k}-G_{k}^{l})^{2},$$
where $\mathcal{G}_{\hat{\theta}}(I^{l})_{k}$ is the predicted volume of class $k$ for the labeled image $I^{l}$, and $G_{k}^{l}$ represents the corresponding ground truth volume of the $k$-th class.
This formulation minimizes the difference between the actual and predicted regional volumes across all labeled images. Then, we formulate the regularization loss by enforcing consistency between the class ratios derived from the segmentation network and those predicted by the regression network for unlabeled images.
Specifically, for each unlabeled image $I^{u}$, the regression network predicts the class ratio $$p^{reg}(u,k,\hat{\theta})=\frac{1}{N}\mathcal{G}_{\hat{\theta}}(I^{u})_{k},$$
which represents the proportion of the image volume occupied by class $k$. Meanwhile, the class ratio derived from the segmentation network's prediction is given by:
$$p^{seg}(u,k,\theta_{1},\hat{\theta})=\frac{1}{N}\sum_{n=1}^{N}\mathcal{S}_{\theta_{1}}(I^{u}(x_{n}), \mathcal{G}_{\hat{\theta}}(I^{u}))_{k}.$$
We consider the class ratio of an image as a categorical distribution. For each unlabeled image, we compute the image-scale Wasserstein distance between the distribution derived from the segmentation output and that predicted by the regression network. This distance is then used to define a regularization loss:
$$\mathcal{L}_{\mathcal{G}}(\theta_{1}, \hat{\theta}) = W(p^{reg}(u,\hat{\theta}, \cdot),  p^{seg}(u,\theta_{1}, \hat{\theta}, \cdot)).$$
The predictions from the regression network serve as effective regularization for the segmentation network through this loss.



\subsection{The proposed weak dataset-scale volume distribution prior}\label{sec4.3}

We consider the volume of each image as a sample point on a low-dimensional manifold within the high-dimensional data space. Consequently, the set of volumes in a dataset can be viewed as realizations from a distribution defined on this manifold, which we refer to as the volume distribution of the dataset. Accordingly, the dataset-scale Wasserstein distance is computed between the volume distribution of labeled dataset and the predicted volume distribution of unlabeled dataset, as shown in Fig. \ref{W}. In Fig. \ref{W}, $\mathcal{V}^{gt}$ and $\mathcal{V}^{pred}$ represent the mapping from volume information to volume distribution for the labeled and unlabeled dataset, respectively.
We introduce a weak dataset-scale volume distribution prior by incorporating this Wasserstein distance as loss function. This assumption is reasonable when the labeled and unlabeled datasets are homologous. 
Specifically, we adopt the Kantorovich-Rubinstein formulation of the $1$-Wasserstein distance $W(\rho^{gt},\rho^{pred})$ as the loss function $\mathcal{L}_{W}$, i.e.,
\begin{align*}
\mathcal{L}_{W} := W(\rho^{gt},\rho^{pred}) &= \frac{1}{M}\mathop{\max}_{\|f\|_{L}\leq M}\{\mathbb{E}_{x\sim  \rho^{gt}}[f(x)]\\
&-\mathbb{E}_{z \sim  \rho^{pred}}[f(z)]\},
\end{align*}
where $\rho^{pred}$ denotes the predicted volume distribution for the unlabeled dataset, obtained from the segmentation outputs of network $\mathcal{S}_{\theta_{1}}$ on the unlabeled images, 
$\rho^{gt}$ denotes the volume distribution for the labeled dataset. The condition $\|f\|_L \leq M$ ensures that $f$ has a Lipschitz constant no greater than $M$.

\setcounter{figure}{2} 
  \begin{figure}[htbp]
\centering         
\begin{overpic}[width=0.7\textwidth]{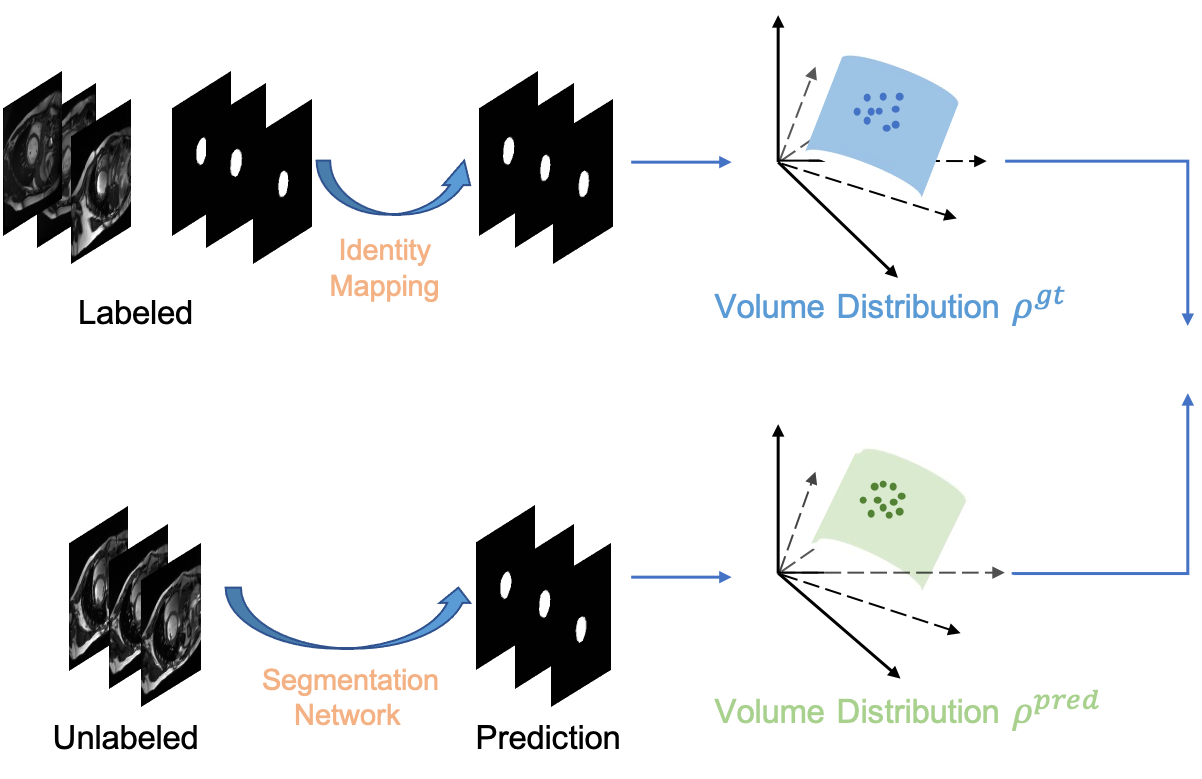}
\put(83.5, 33.5){ \footnotesize$W(\rho^{gt},\rho^{pred})$}
\put(53, 53.4){\footnotesize $\mathcal{V}^{gt}$}
\put(51.5, 18.8){\footnotesize $\mathcal{V}^{pred}$}
\end{overpic}

\caption{Illustration of the construction of volume distribution weak prior. $\mathcal{V}^{gt}$ and $\mathcal{V}^{pred}$ represent the mapping from volume information to volume distribution, with $\mathcal{V}^{gt}$ for labeled and $\mathcal{V}^{pred}$ for unlabeled datasets.}\label{W}
\end{figure}

 \subsection{ Overall Loss Function}
The overall loss function for the proposed semi-supervised segmentation network combines the segmentation loss \( \mathcal{L}_{\mathcal{S}}(\theta_{1}) \), the regularization loss \( \mathcal{L}_{\mathcal{G}}(\theta_{1}, \hat{\theta}) \), and the weak volume prior-based loss. It is formulated as follows:
$$\mathcal{L}=\mathcal{L}_{\mathcal{S}}(\theta_{1})+\alpha\mathcal{L}_{\mathcal{G}}(\theta_{1}, \hat{\theta})+ \beta \mathcal{L}_{W},$$
where $\alpha$ and $\beta$ are weighting coefficients that control the influence of each loss term, both set to $0.01$.
This leads to the following optimization problem:
\begin{align*}
\mathop{\min}_{\theta_{1}} & \big\{ \mathcal{L}_{\mathcal{S}}(\theta_{1}) + \alpha \mathcal{L}_{\mathcal{G}}(\theta_{1}, \hat{\theta}) \\
&  + \frac{\beta}{M} \mathop{\max}_{\|f\|_{L} \leq M} \left\{ \mathbb{E}_{x \sim \rho^{gt}}[f(x)] - \mathbb{E}_{z \sim \rho^{pred}}[f(z)] \right\} \big\}.
\end{align*}
We approximate the function $f$ using a network $\mathcal{F}_{\theta_{2}}$ with parameters $\theta_{2}$. The set of networks with a Lipschitz constant less than or equal to $M$ is denoted as $\mathbb{F}_{M}$.
To approximate the expectation of the random variables, we compute the mean over a sample set. As a result, we can reformulate the problem as follows:
\begin{equation*}
\begin{aligned}
\mathop{\min}_{\theta_{1}} \Bigg\{ & \mathcal{L}_{\mathcal{S}}(\theta_{1}) 
+ \alpha \mathcal{L}_{\mathcal{G}}(\theta_{1}, \hat{\theta}) \\
& + \frac{\beta}{M} \mathop{\max}_{\theta_{2}, \mathcal{F}_{\theta_{2}}\in\mathbb{F}_{M} } \Big\{ 
\frac{1}{L_{1}} \sum_{\mathbf{V}_{gt}^{l}\in \mathbb{V}_{gt}} \mathcal{F}_{\theta_{2}}(\mathbf{V}_{gt}^{l})  \\
&- \frac{1}{U_{1}} \sum_{I^{u} \in \mathbb{U}} \mathcal{F}_{\theta_{2}}\Big( \sum_{n=1}^{N} 
\mathcal{S}_{\theta_{1}}\big(I^{u}(x_{n}), \mathcal{G}_{\hat{\theta}}(I^{u})\big) \Big) \Big\} \Bigg\},
\end{aligned}
\end{equation*}
where $L_{1}$ stands for the number of samples in the volume set $\mathbb{V}_{gt}$ of the labeled images, and $U_{1}$ denotes the number of samples in $\mathbb{U}$. 
An alternating optimization strategy is adopted to approximately solve the above problem. We define
\begin{equation*}
\begin{aligned}
\mathcal{L}_{1}(\theta_{1}): = & \mathcal{L}_{\mathcal{S}}(\theta_{1})+\alpha \mathcal{L}_{\mathcal{G}}(\theta_{1},\hat{\theta}) \\
&-\frac{\beta}{MU_{1}}
\sum_{I^{u}\in\mathbb{U}}\mathcal{F}_{\theta_{2}}\Big( \sum_{n=1}^{N} 
\mathcal{S}_{\theta_{1}}\big(I^{u}(x_{n}), \mathcal{G}_{\hat{\theta}}(I^{u})\big)\Big),
\end{aligned}
\end{equation*}
\begin{equation*}
\begin{aligned}
\mathcal{L}_{2}(\theta_{2}) &:= \frac{1}{U_{1}}\sum_{I^{u}\in\mathbb{U}}\mathcal{F}_{\theta_{2}}\Big( \sum_{n=1}^{N} 
\mathcal{S}_{\theta_{1}}\big(I^{u}(x_{n}), \mathcal{G}_{\hat{\theta}}(I^{u})\big)\Big)\\
&- \frac{1}{L_{1}}\sum_{\mathbf{V}_{gt}^{l}\in\mathbb{V}_{gt}}\mathcal{F}_{\theta_{2}}(\mathbf{V}_{gt}^{l}),
\end{aligned}
\end{equation*}
and train the entire network by alternately minimizing the following two subproblems:
\begin{equation*}\label{dncnn}
\left\{
\begin{aligned}
&\mathop{\min}_{\theta_{1}}\mathcal{L}_{1}(\theta_{1})
,\\
&\mathop{\min}_{\theta_{2}}\mathcal{L}_{2}(\theta_{2}).
\end{aligned}
\right.
\end{equation*}
First, we fix $\mathcal{F}_{\theta_{2}}$ and train the segmentation network by minimizing $\mathcal{L}_{1}(\theta_{1})$. Then, we fix the segmentation network, and minimize $\mathcal{L}_{2}(\theta_{2})$ with respect to $\theta_{2}$. These two steps are alternated throughout the training process.

\section{Numerical experiments}
\label{experiment}
\noindent 
\subsection{Setup}
\textbf{Data.} 
We conduct experiments on the Automated Cardiac Diagnosis Challenge (ACDC) segmentation dataset \citep{ACDC}, the thigh muscle MR images, and the Prostate MR Image Segmentation (PROMISE12) dataset \citep{promise}.

\textbf{Implementation details.}
We implement the semi-supervised segmentation framework in Python with PyTorch.
For the regression network, ResNeXt-101 \citep{regressor} is employed as the backbone architecture, with the squared $L_{2}$ norm  serving as optimization objective. The network is trained for 200 epochs using stochastic gradient descent. 
As is common in most semi-supervised segmentation studies, we adopt the U-Net \citep{unet} as the backbone segmentation network. Specifically, we enhance U-Net by incorporating a softmax layer with volume-preservation and TD spatial regularization priors as its final layer. We employ the Adam optimizer for training, setting the initial learning rate to $1 \times 10^{-3}$. If the validation Dice Similarity Coefficient (DSC) does not improve within 20 epochs, the learning rate is halved. Training is performed over 100 epochs with a batch size of 4. The network $\mathcal{F}_{\theta_{2}}$ is selected as a simple linear layer network with weight clipping.

\textbf{Comparative Methods.} 
We evaluate our proposed method by comparing it against several baseline methods and SOTA semi-supervised methods.
We train U-Net \citep{unet}, the backbone segmentation network, using labeled images and their corresponding pixel-level annotations.
Following this, according to the standard self-training semi-supervised strategy, predictions for the unlabeled images are generated and integrated into the training set as pseudo-labels. This approach is referred to as the self-training (S-T) method.
Additionally, we choose the relative semi-supervised segmentation methods, such as the curriculum semi-supervised (C-S) method \citep{curr}, the consistency learning method UA-MT \citep{utma}, EVIL \citep{chen2024evidence} and MC-Net+ \citep{mc}.
The comparative results between our proposed VP-Net and these methods are presented in Tables \ref{acdc}, \ref{thigh}, and \ref{promise}, with top-performing values highlighted in bold.

To further validate the effectiveness associated with the strong volume prior derived from the variational model, we train a variant of VP-Net, VP-Net\textsubscript{gt}, which uses ground-truth volumes instead of the regression network output for unlabeled images. 
Tables \ref{acdc}, \ref{thigh}, and \ref{promise} summarize the comparison between VP-Net\textsubscript{gt} and competing methods, where superior results are underlined.
These results demonstrate the effectiveness of incorporating the strong volume prior to improve semi-supervised segmentation networks.

\textbf{Evaluation.} We evaluate the segmentation methods using quantitative metrics such as Average Surface Distance (ASD), Dice Similarity Coefficient (DSC), Jaccard, and 95\% Hausdorff Distance (95HD). Additionally, we offer visual comparisons of different segmentation methods for qualitative evaluation.

\subsection{Results on ACDC2017}

\begin{figure*}[ht]
    \centering
    \subfloat{\includegraphics[width=0.122\textwidth]{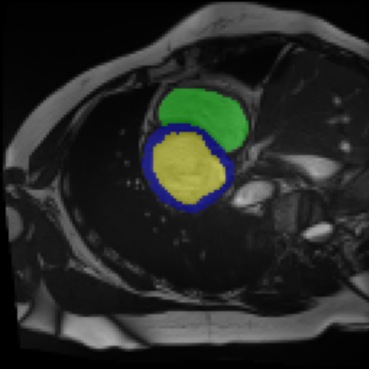}}  \hspace{-0.007\textwidth}
    \subfloat{\includegraphics[width=0.122\textwidth]{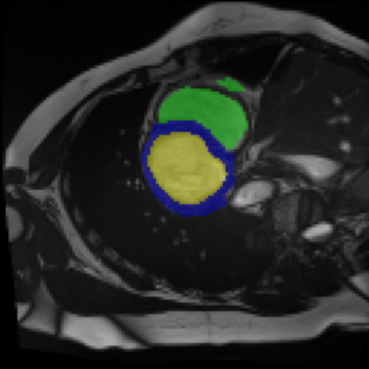}} \hspace{-0.007\textwidth}
    \subfloat{\includegraphics[width=0.122\textwidth]{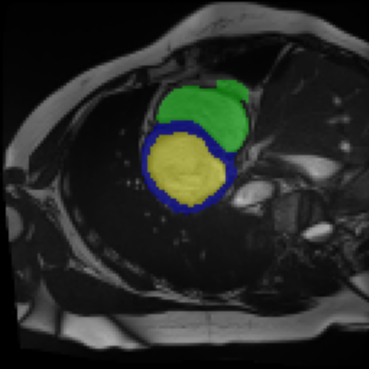}} \hspace{-0.007\textwidth}
    \subfloat{\includegraphics[width=0.122\textwidth]{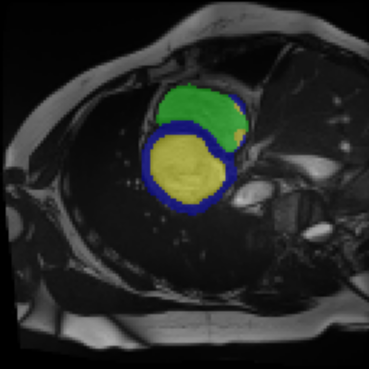}} \hspace{-0.007\textwidth}
    \subfloat{\includegraphics[width=0.122\textwidth]{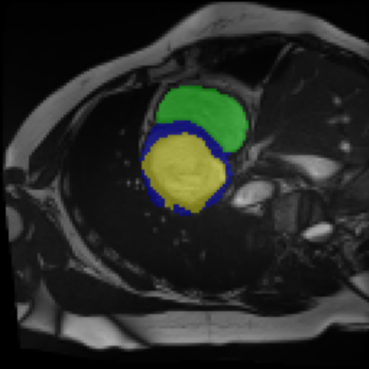}} \hspace{-0.007\textwidth}
    \subfloat{\includegraphics[width=0.122\textwidth]{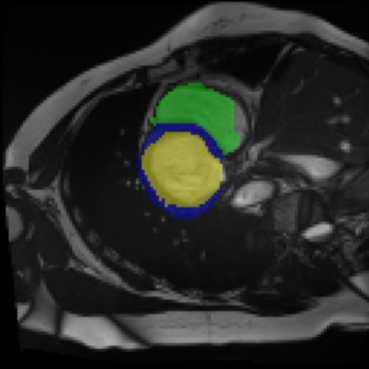}} \hspace{-0.007\textwidth}
        \subfloat{\includegraphics[width=0.122\textwidth]{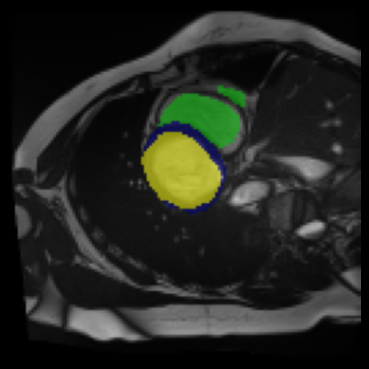}}\hspace{-0.007\textwidth}
    \subfloat{\includegraphics[width=0.122\textwidth]{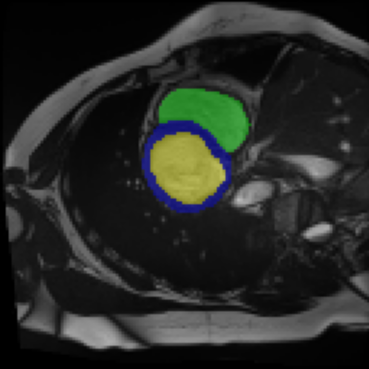}} \hspace{-0.007\textwidth}\\
    \vspace{0.4pt}
    \subfloat{\includegraphics[width=0.122\textwidth]{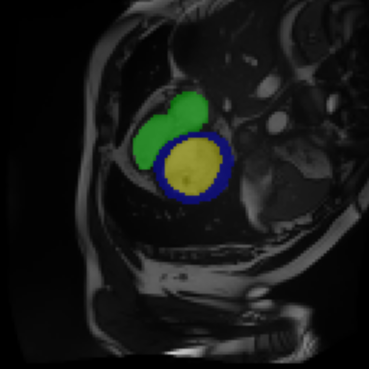}} \hspace{-0.007\textwidth}
    \subfloat{\includegraphics[width=0.122\textwidth]{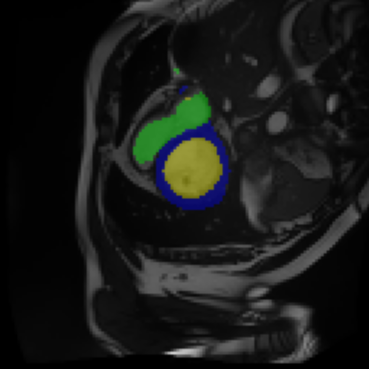}} \hspace{-0.007\textwidth}
    \subfloat{\includegraphics[width=0.122\textwidth]{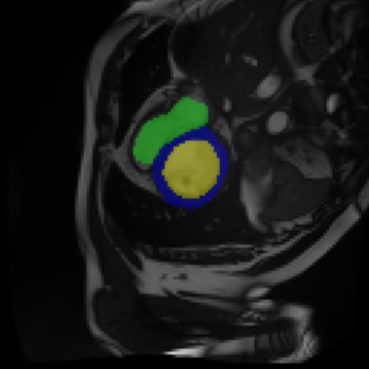}}\hspace{-0.007\textwidth}
    \subfloat{\includegraphics[width=0.122\textwidth]{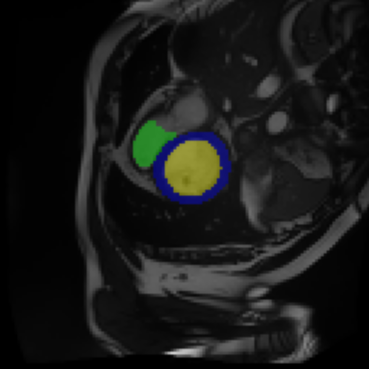}} \hspace{-0.007\textwidth}
    \subfloat{\includegraphics[width=0.122\textwidth]{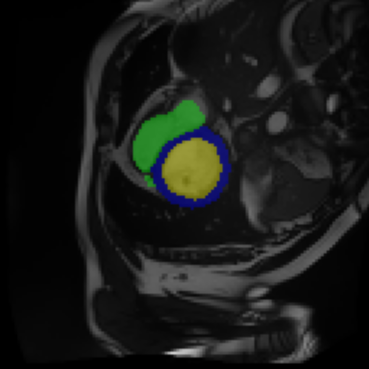}} \hspace{-0.007\textwidth}
    \subfloat{\includegraphics[width=0.122\textwidth]{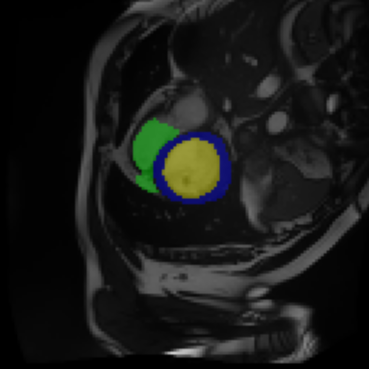}}\hspace{-0.007\textwidth}
    \subfloat{\includegraphics[width=0.122\textwidth]{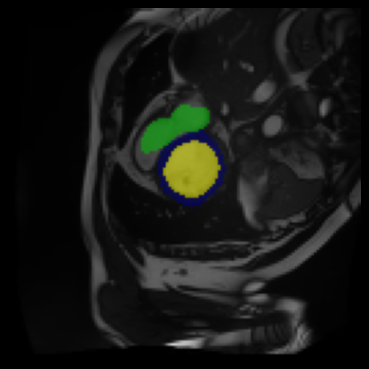}}\hspace{-0.007\textwidth}
    \subfloat{\includegraphics[width=0.122\textwidth]{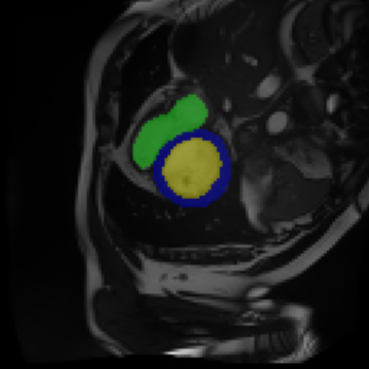}}\hspace{-0.007\textwidth}\\
  \vspace{0.4pt}
  \setcounter{subfigure}{0}
        \subfloat[{ G-T}]{\includegraphics[width=0.122\textwidth]{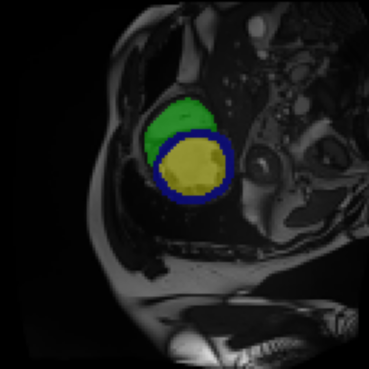}} \hspace{-0.007\textwidth}
    \subfloat[{U-Net}]{\includegraphics[width=0.122\textwidth]{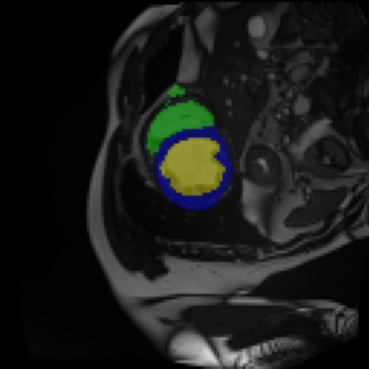}}\hspace{-0.007\textwidth}
    \subfloat[{S-T}]{\includegraphics[width=0.122\textwidth]{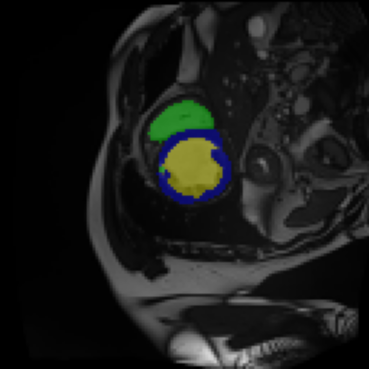}} \hspace{-0.007\textwidth}
    \subfloat[{C-S}]{\includegraphics[width=0.122\textwidth]{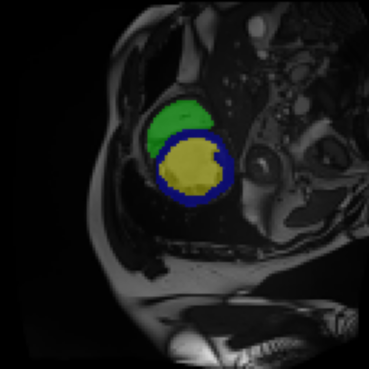}}\hspace{-0.007\textwidth}
    \subfloat[{UA-MT}]{\includegraphics[width=0.122\textwidth]{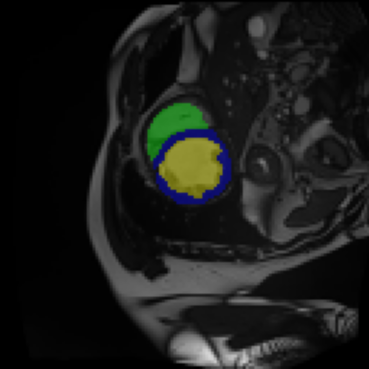}} \hspace{-0.007\textwidth}
    \subfloat[\tiny{MC-Net+}]{\includegraphics[width=0.122\textwidth]{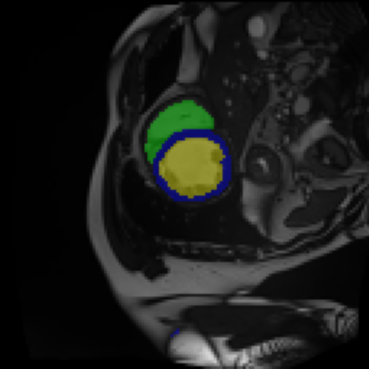}} \hspace{-0.007\textwidth}
        \subfloat[EVIL]{\includegraphics[width=0.122\textwidth]{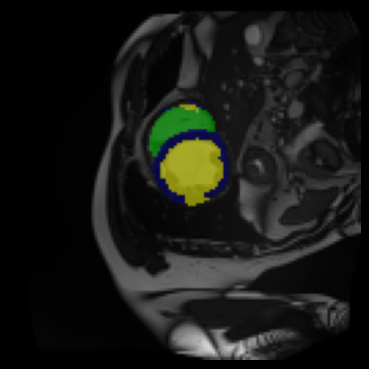}} \hspace{-0.007\textwidth}
    \subfloat[{VP-Net}]{\includegraphics[width=0.122\textwidth]{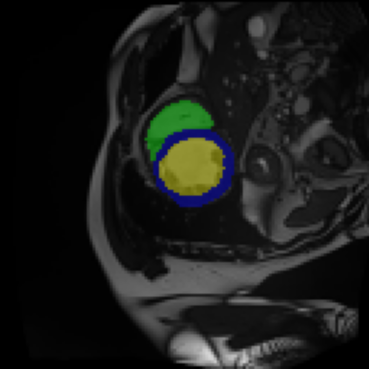}} \hspace{0.000\textwidth}

        \caption{Visual comparison of different segmentation approaches on the ACDC2017 dataset. G-T: Ground Truth.
}\label{acdcimg}
\end{figure*}

The ACDC2017 dataset \citep{ACDC} consists of 200 short-axis cine-MRI scans. The objective is to delineate four regions of interest: the left ventricle cavity (LV), the left ventricle myocardium (Myo), the right ventricular cavity (RV), and the background (BG). We extract two-dimensional slices from 3D MRI scans for experimentation. The images were resized to $128\times 128$ without any additional preprocessing.
We use 140 labeled exams for training, with the remaining 60 used for validation in the experiments. 
The training set consists of $n$ fully labeled images, while pixel-wise annotations are not available for the remaining $140 -n$ images.
The comparison results are presented in Table \ref{acdc}. Compared to supervised method U-Net \citep{unet}, all semi-supervised methods achieve better results in terms of both quantitative metrics.
Fig. \ref{acdcimg} provides a visual comparison of the segmentation results across different segmentation models. 
As observed, the segmentation results from other methods exhibit missing regions, whereas our proposed method produces more complete and coherent segmentation areas. This demonstrates the effectiveness of incorporating the volume-preservation prior into the network.

\begin{table}[htbp] 
\centering 
\caption{Quantitative evaluation on the ACDC2017 dataset. Best results for VP-Net against other methods are shown in bold, while those for VP-Net\textsubscript{gt} are underlined.
}\label{acdc}
\resizebox{1.0\textwidth}{!}{
\begin{tabular}{p{3.6cm}p{1.3cm}p{1.7cm}p{0.2cm}p{1.8cm}p{2.4cm}p{2.4cm}p{2.4cm}}
\hline
\multirow{2}{*}{Method} &\multicolumn{2}{l}{ \# of scans} &  &\multicolumn{4}{l}{Metrics} \\
\cline{2-3} \cline{5-8}
 &Labeled & Unlabeled & & Dice (\%)$\uparrow$ & Jaccard (\%)$\uparrow$  &ASD(voxel)$\downarrow$ &95HD(voxel)$\downarrow$ \\
\hline
U-Net & 60  & 80 & &70.70& 64.24& 6.91& 12.20  \\
U-Net & 100 & 40 & & 76.92 &72.47 &5.34 &  6.54 \\
U-Net &120 & 20 & &  79.90& 75.04& 4.30 &  5.50\\
\hline
 Self-training& & & & 70.45&  64.41&  6.45&  8.56 \\
Curriculum & & & &73.12 & 64.63&  5.16&  10.36\\
UA-MT& & & &71.52 & 64.38& 6.62&  9.78\\
MC-Net+ & 60& 80 &  &72.95 &  66.85& 5.50&  7.64  \\
EVIL & & & & 72.66 & 62.62 & 6.53 & 13.37 \\
\textbf{VP-Net(Ours)} & & &&\textbf{76.73}& \textbf{69.69}& \textbf{3.57}&  \textbf{5.82} \\
\textbf{VP-Net\textsubscript{gt}(Ours)} & & & &\underline{79.93} & \underline{75.79} & \underline{4.89} &  \underline{7.52} \\
\hline
 Self-training& && & 78.16  &  73.03& 4.62 & 6.33\\
Curriculum & & & & 77.24& 72.16 &  4.57 &  5.89\\
UA-MT& & & &77.14 & 73.06  &  5.33 &  6.55\\
MC-Net+ & 100 & 40 & & 79.61& 75.40& 4.94&  6.73\\
EVIL & & & & 79.82 & 69.87 & 6.09 & 9.92\\
\textbf{VP-Net(Ours)} & & & & \textbf{81.40}&  \textbf{75.42}& \textbf{3.78}&  \textbf{5.74}\\
\textbf{VP-Net\textsubscript{gt}(Ours)} & & & & \underline{82.22} &  \underline{79.34} &  \underline{3.52}&  \underline{4.62}\\
\hline
 Self-training& & & &79.85 &  73.64&  3.30&  4.73\\
Curriculum & & & &81.05 & 76.37&  4.29&  5.83\\
UA-MT& & & &80.10 & 74.61& 3.94 & 5.10\\
MC-Net+ & 120 & 20 & & 80.99& 75.75& \textbf{4.08} &  \textbf{5.44}\\
EVIL & & & &81.90  & 75.97 & 4.50 & 6.30 \\
\textbf{VP-Net(Ours)} & & & &\textbf{82.51 }& \textbf{76.58}& 4.09&  6.13\\
\textbf{VP-Net\textsubscript{gt}(Ours)} & & & &\underline{83.66} & \underline{81.19}&  \underline{3.20}&  \underline{4.30}\\
\hline
\end{tabular}
}
\end{table}

\subsection{Results for thigh muscle MR images}
\begin{figure*}[ht]
    \centering
    \begin{minipage}[b]{0.13\textwidth}
        \centering
        \includegraphics[width=\textwidth]{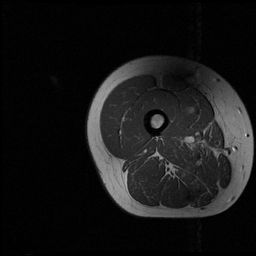}
    \end{minipage}
    \hspace{0.5cm}
    \begin{minipage}[b]{0.13\textwidth}
        \centering
        \includegraphics[width=\textwidth]{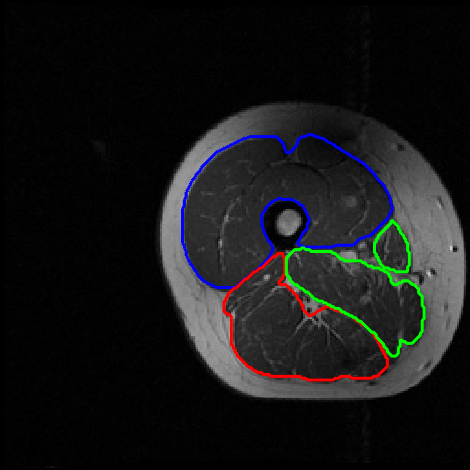}
    \end{minipage}
    \caption{Example T1-weighted thigh MR image. Left: original image. Right: expert-annotated segmentation with distinct colors indicating quadriceps (blue), hamstrings (red), and other muscles (green).}\label{MR_example}
\end{figure*}

\begin{figure*}[htbp]
    \centering
    \subfloat{\includegraphics[width=0.126\textwidth]{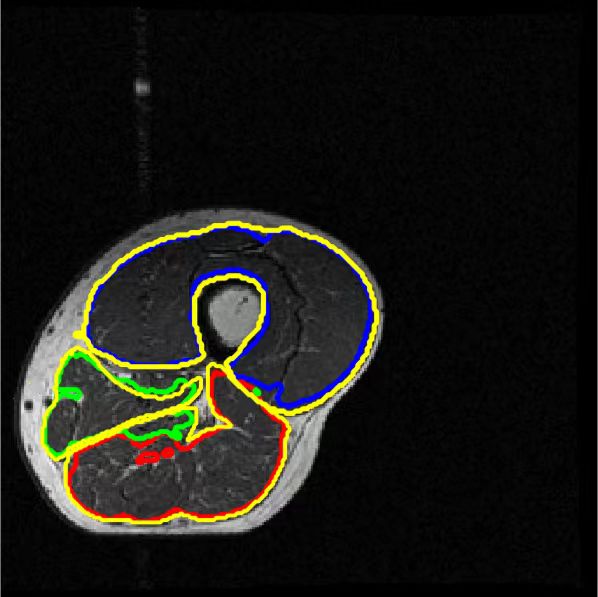}} \hspace{0.01\textwidth}
    \subfloat{\includegraphics[width=0.126\textwidth]{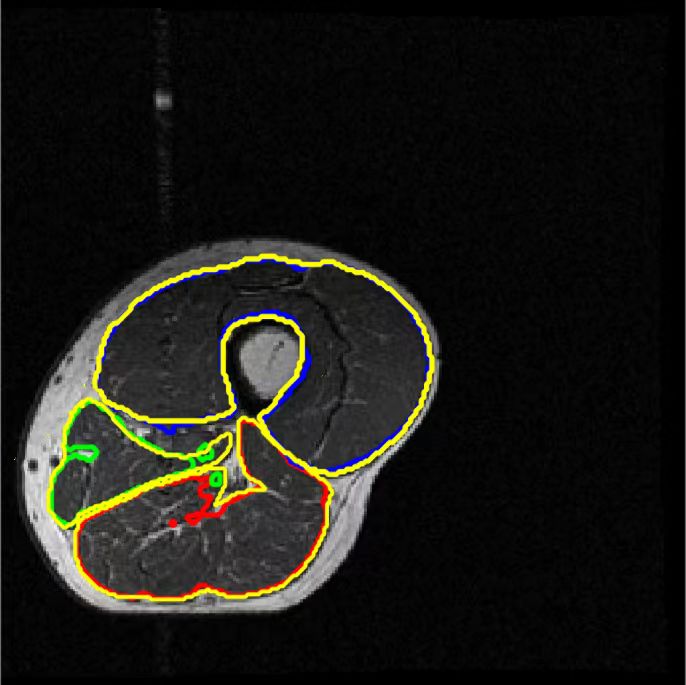}}\hspace{0.01\textwidth}
    \subfloat{\includegraphics[width=0.126\textwidth]{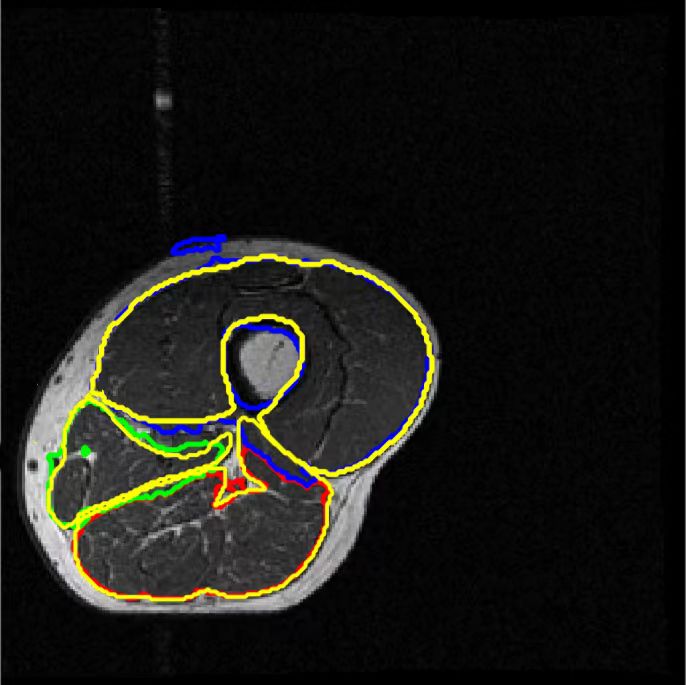}}\hspace{0.01\textwidth}
    \subfloat{\includegraphics[width=0.126\textwidth]{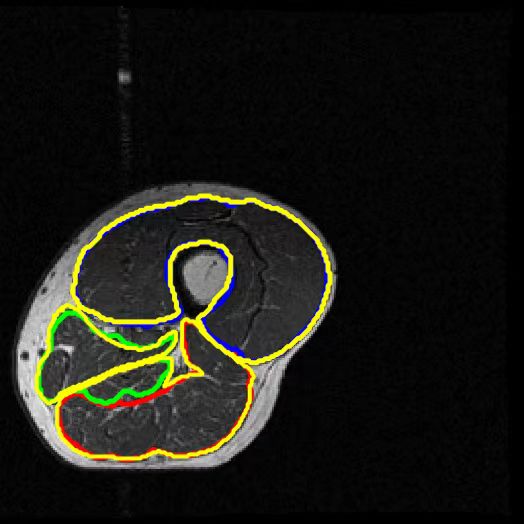}}\hspace{0.01\textwidth}
    \subfloat{\includegraphics[width=0.126\textwidth]{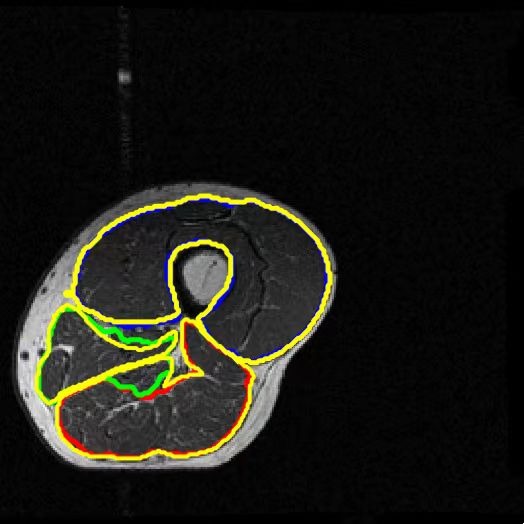}} \hspace{0.01\textwidth}    
     \subfloat{\includegraphics[width=0.126\textwidth]{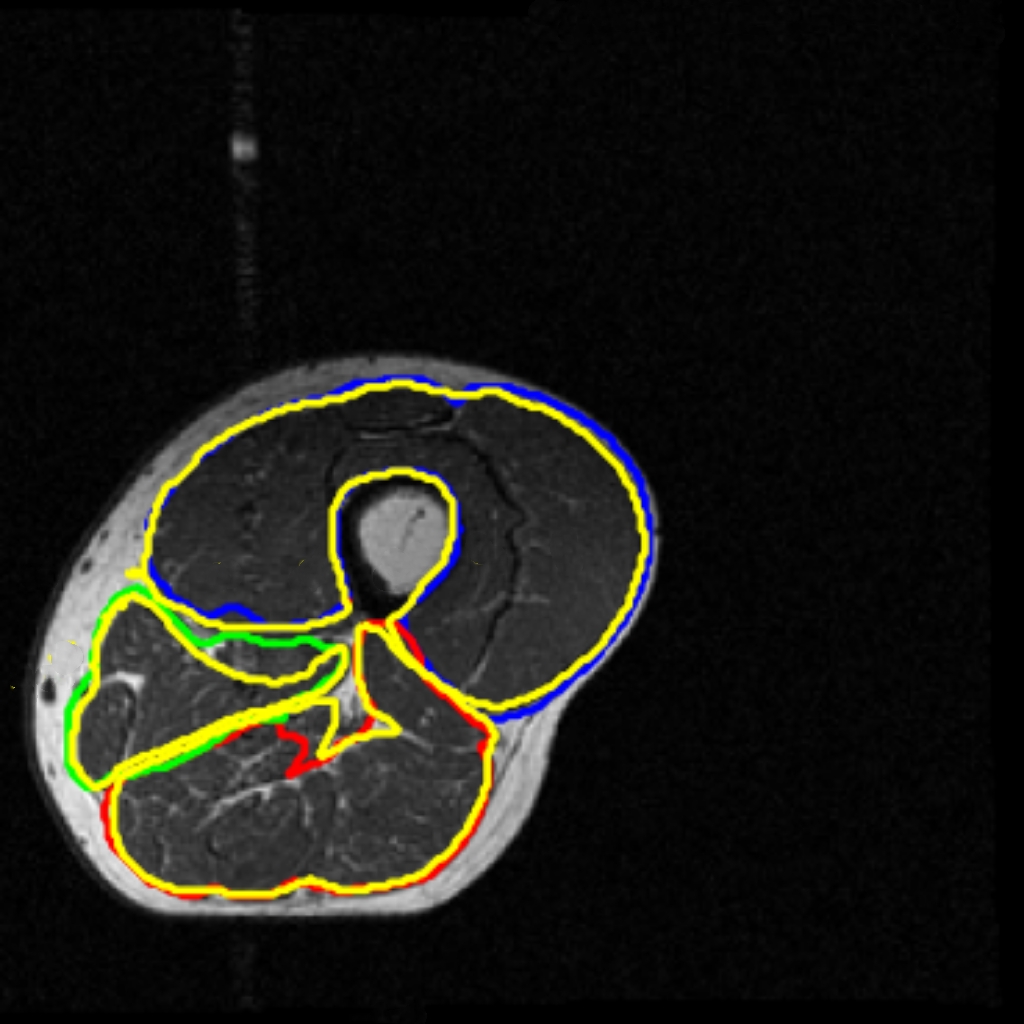}} \hspace{0.01\textwidth}   \subfloat{\includegraphics[width=0.126\textwidth]{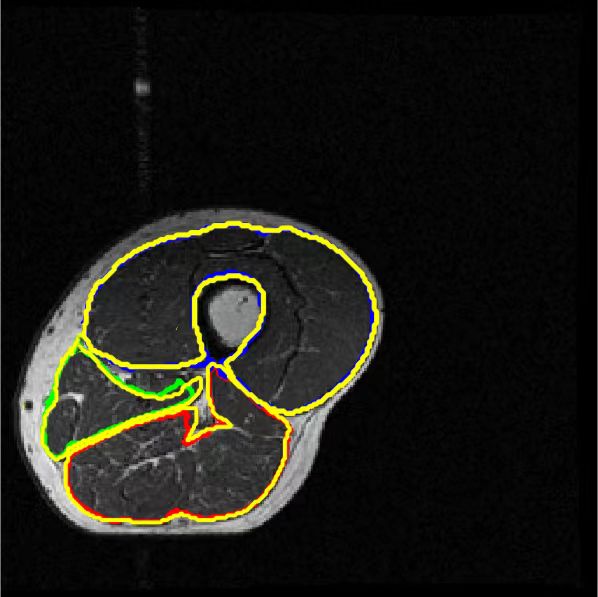}}\\
 \vspace{0.4pt}
    \subfloat{\includegraphics[width=0.126\textwidth]{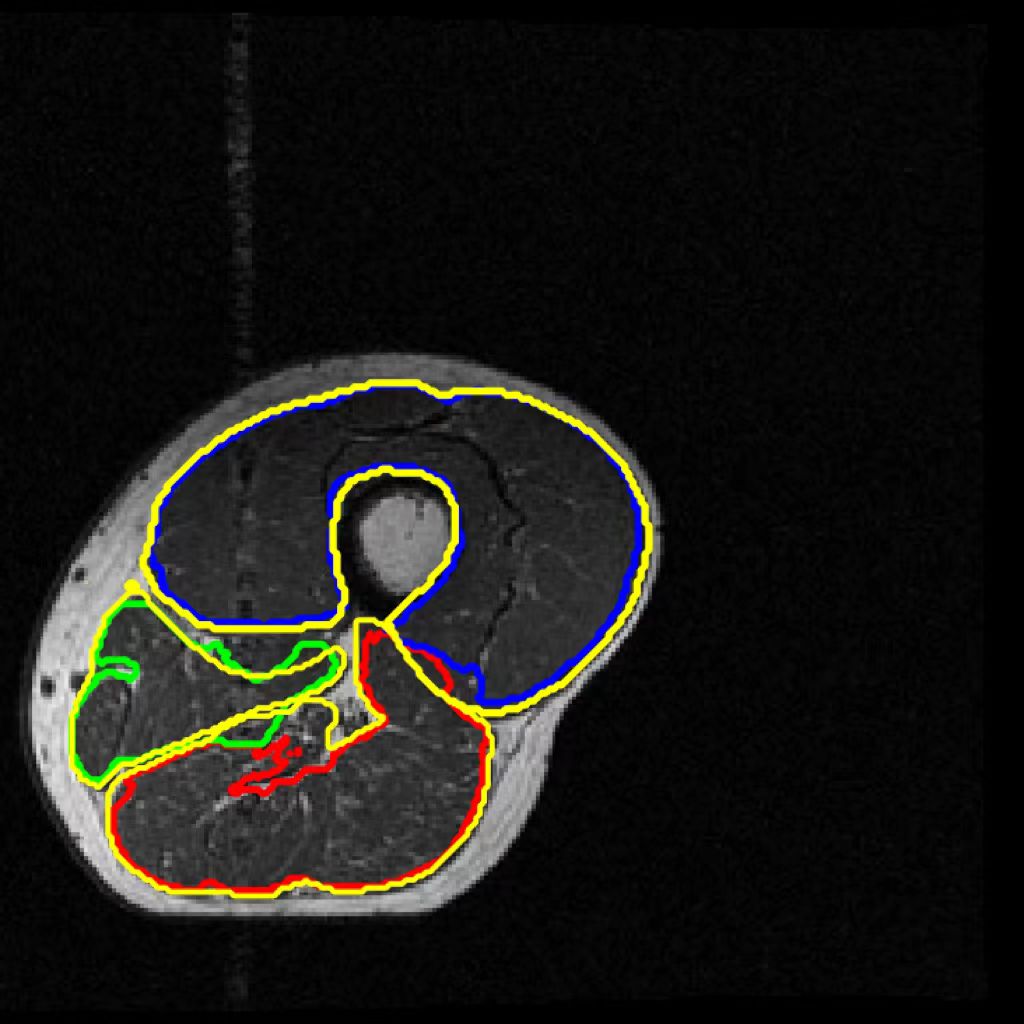}} \hspace{0.01\textwidth}
    \subfloat{\includegraphics[width=0.126\textwidth]{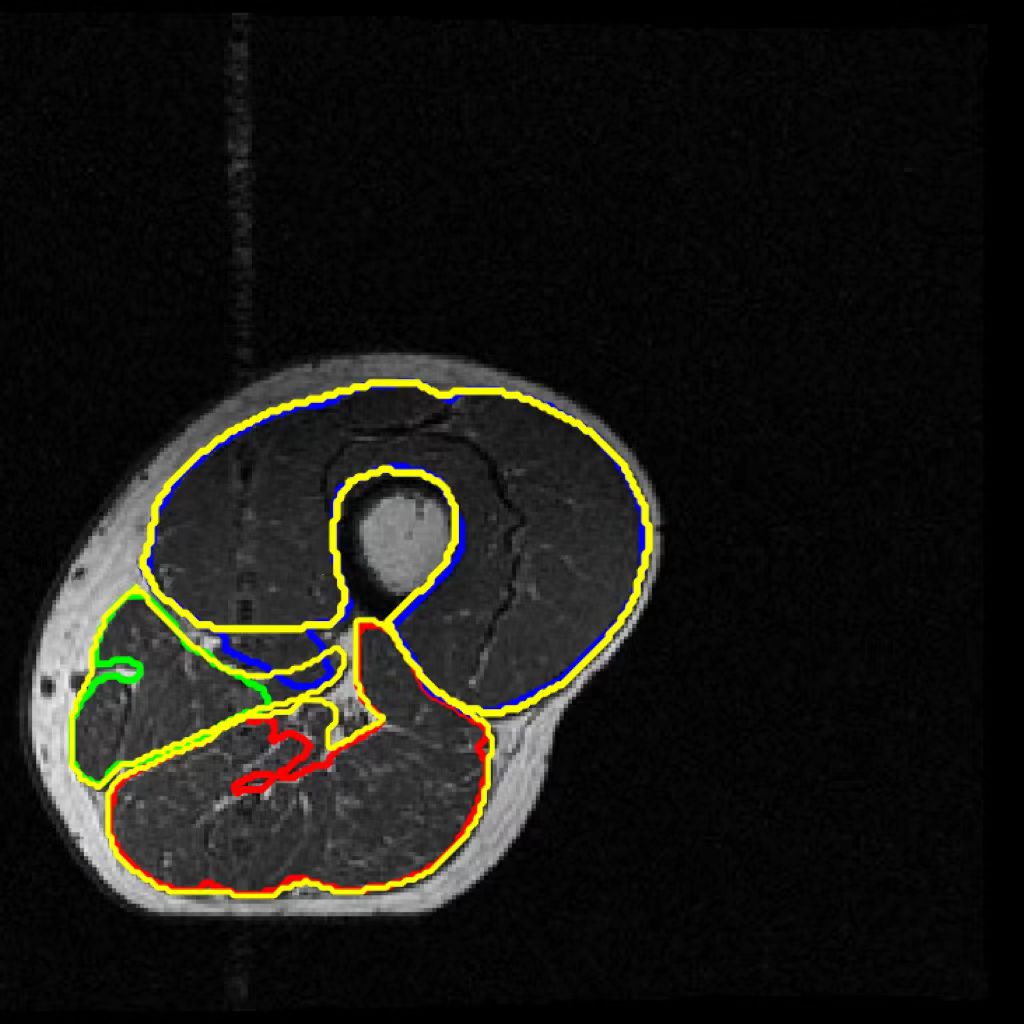}}\hspace{0.01\textwidth}
    \subfloat{\includegraphics[width=0.126\textwidth]{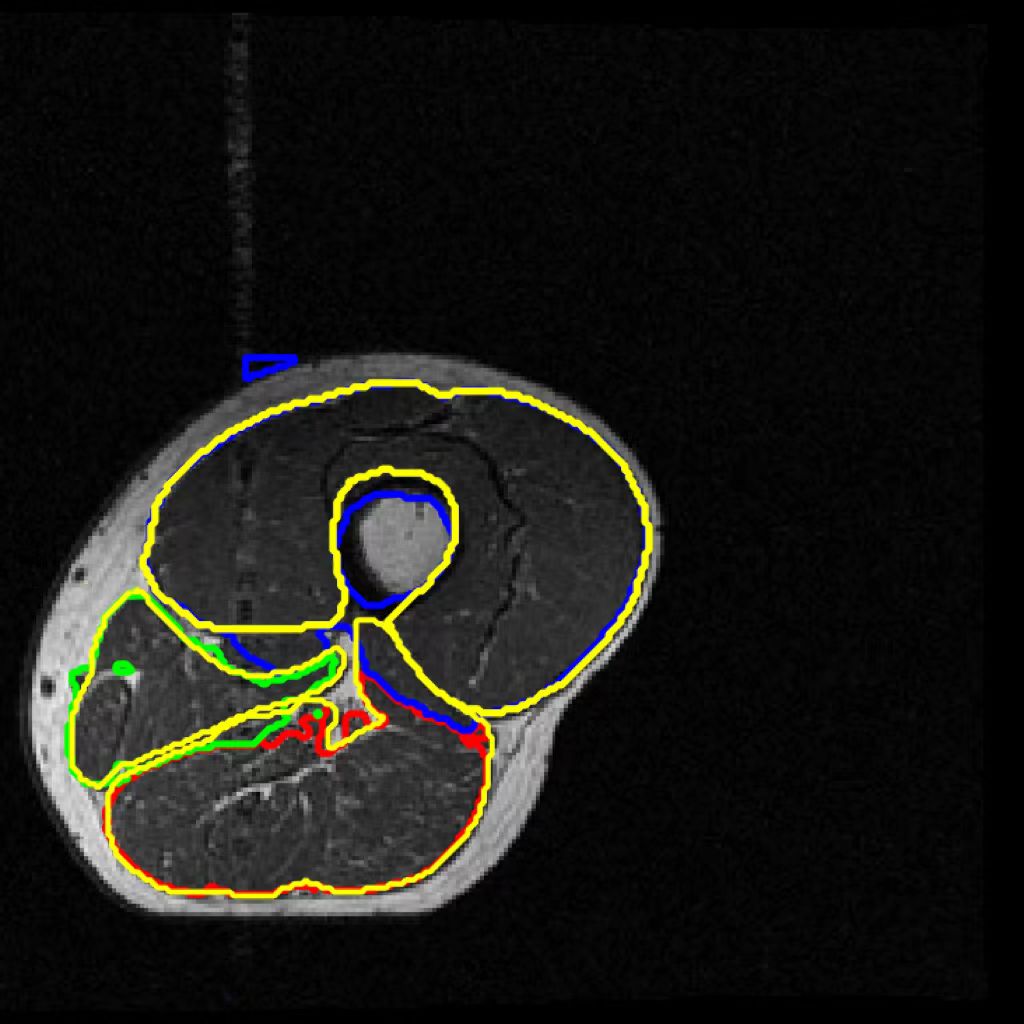}}\hspace{0.01\textwidth}
    \subfloat{\includegraphics[width=0.126\textwidth]{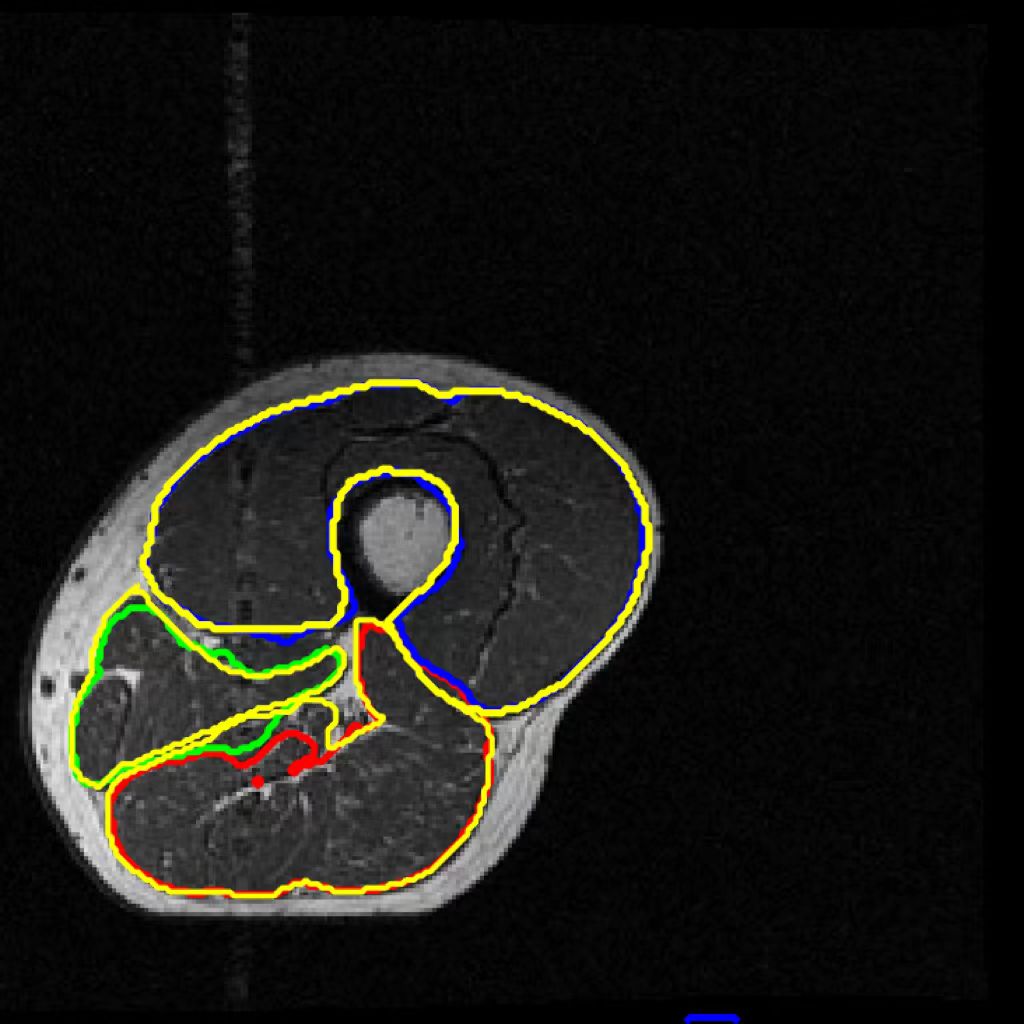}}\hspace{0.01\textwidth}
    \subfloat{\includegraphics[width=0.126\textwidth]{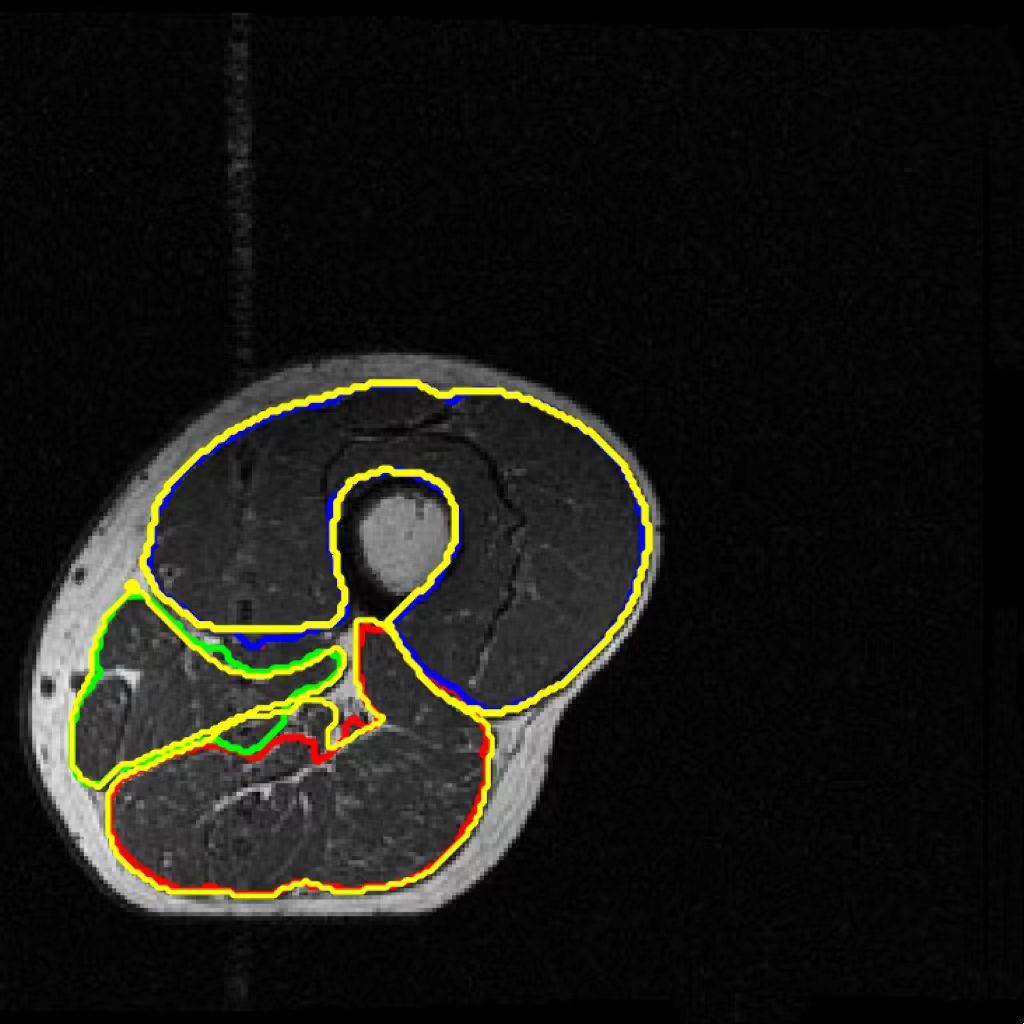}}\hspace{0.01\textwidth}
        \subfloat{\includegraphics[width=0.126\textwidth]{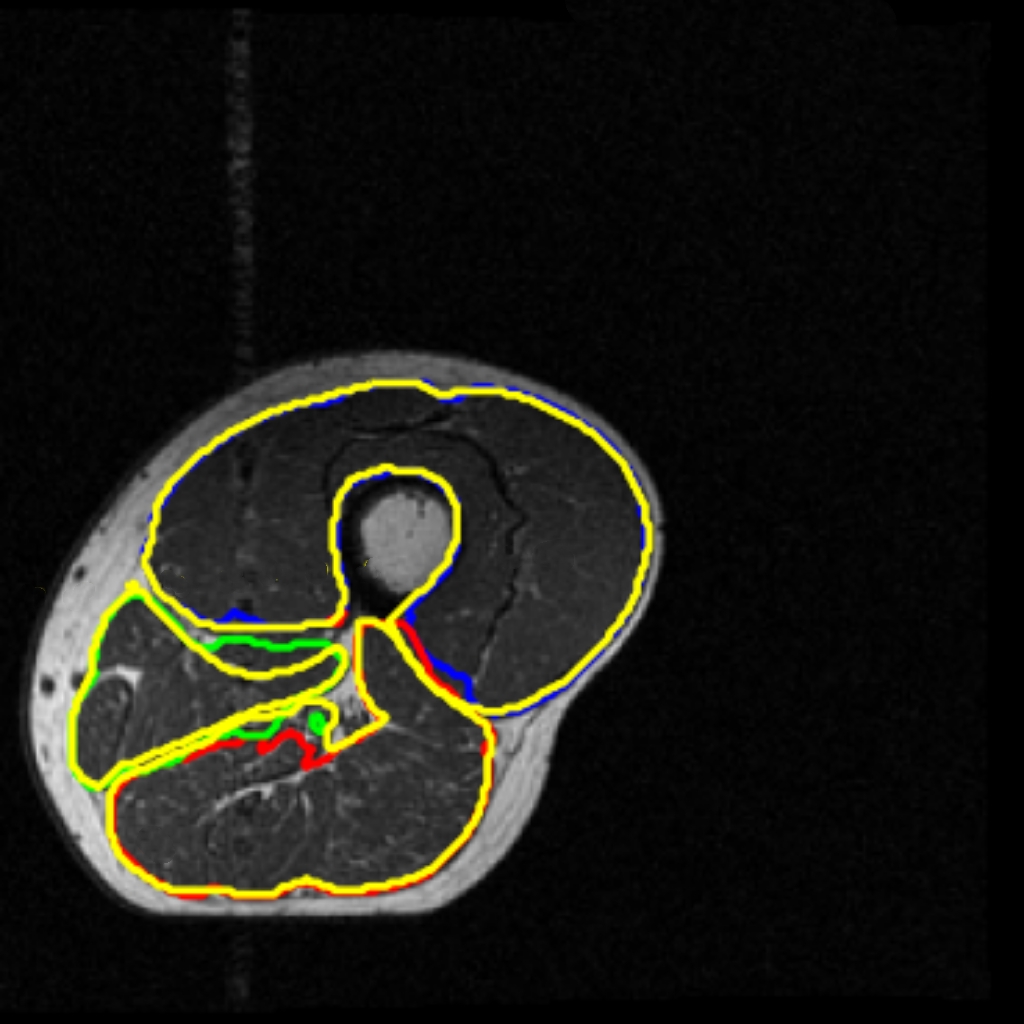}}\hspace{0.01\textwidth}
    \subfloat{\includegraphics[width=0.126\textwidth]{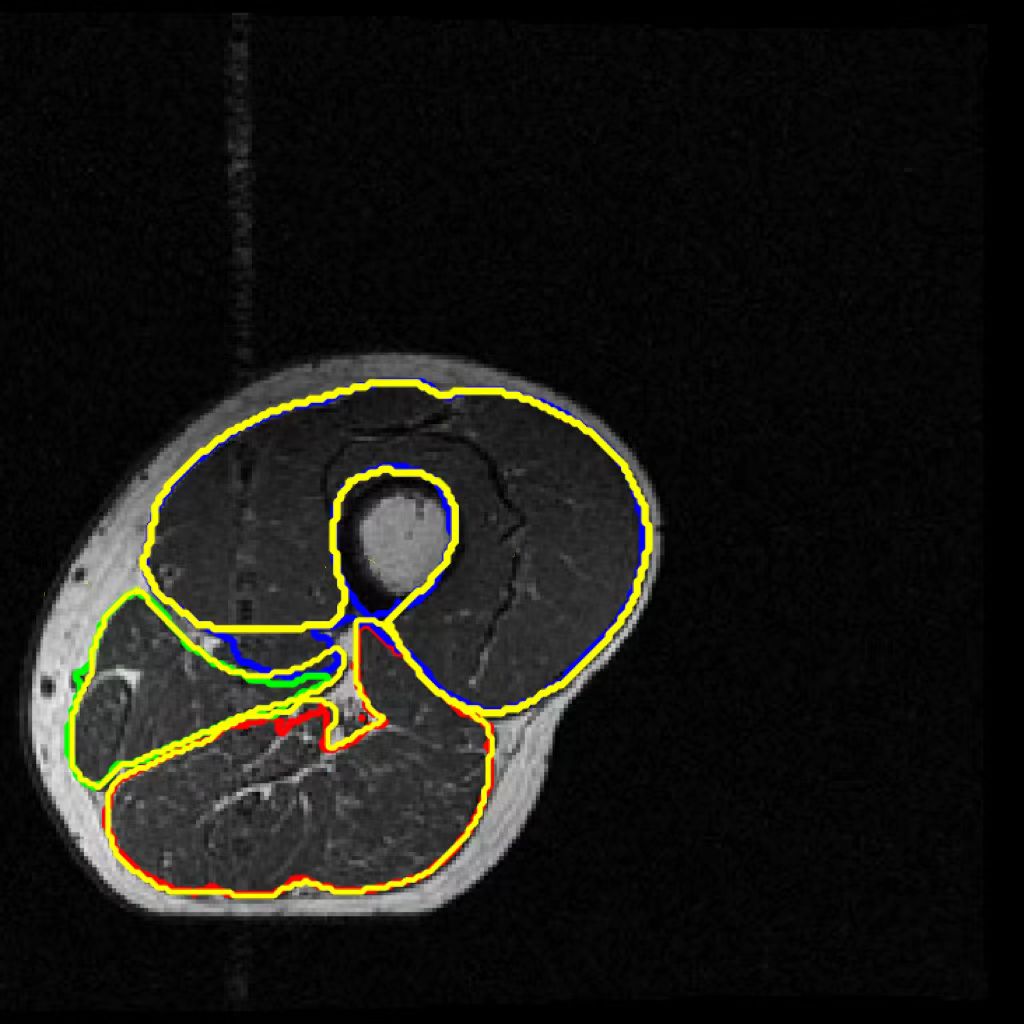}} \\
 \vspace{0.4pt}
  \setcounter{subfigure}{0}
\subfloat[U-Net]{\includegraphics[width=0.126\textwidth]{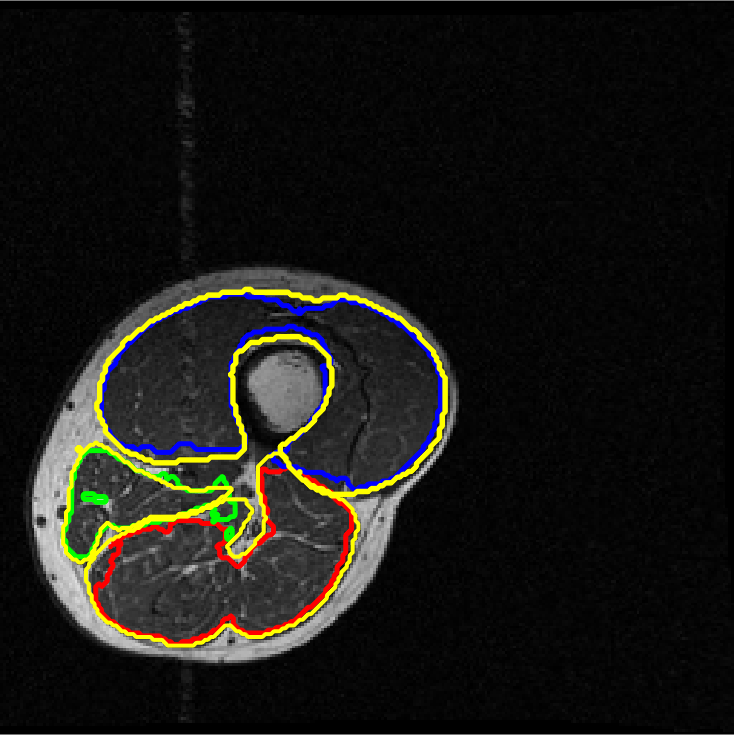}} \hspace{0.01\textwidth}
\subfloat[{S-T}]{\includegraphics[width=0.126\textwidth]{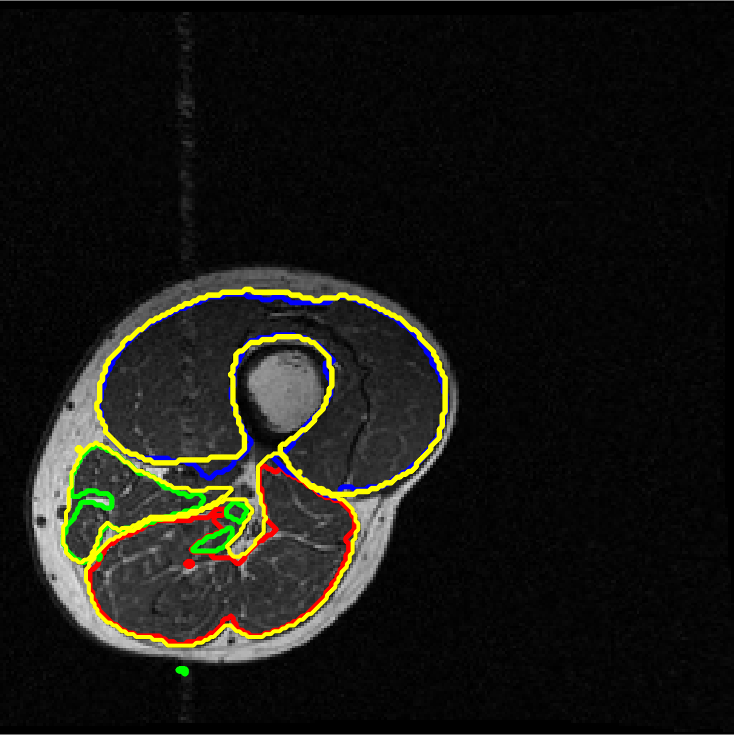}} \hspace{0.01\textwidth}
\subfloat[{C-S}]{\includegraphics[width=0.126\textwidth]{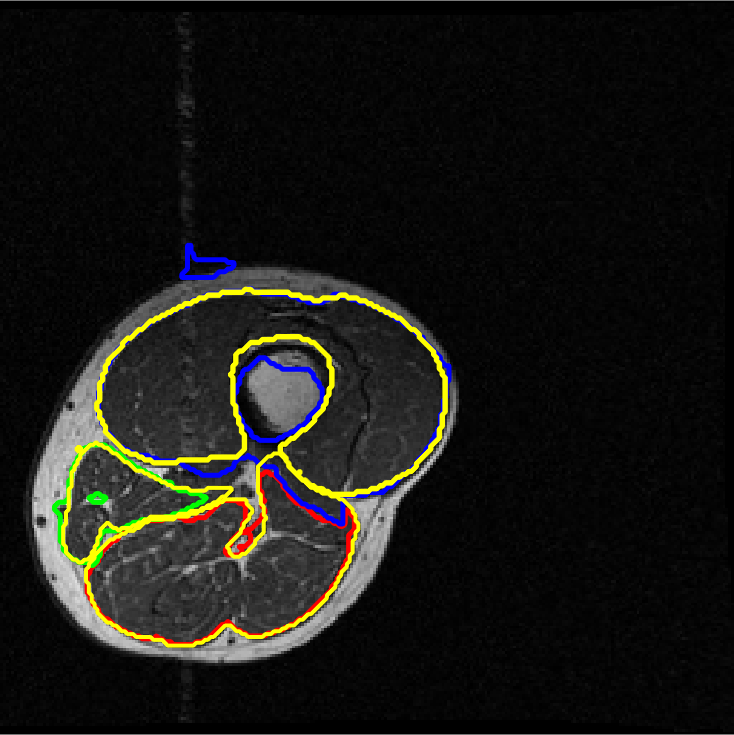}} \hspace{0.01\textwidth}
\subfloat[{UA-MT}]{\includegraphics[width=0.126\textwidth]{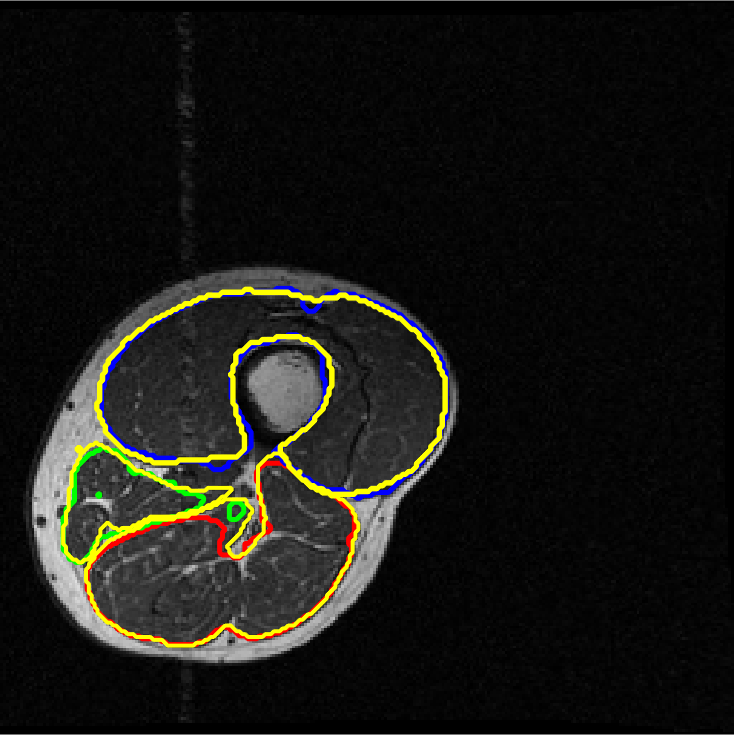}} \hspace{0.01\textwidth}
\subfloat[{MC-Net+}]{\includegraphics[width=0.126\textwidth]{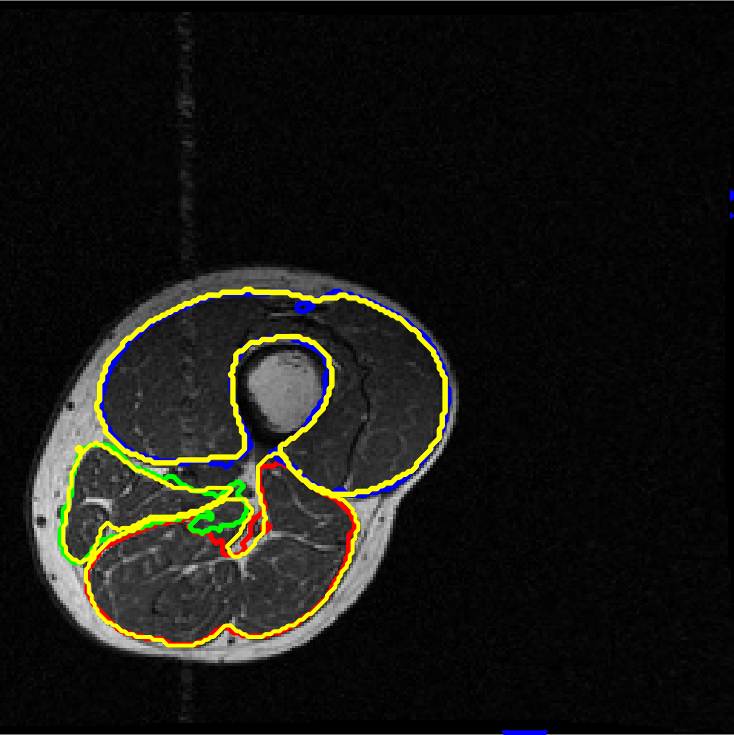}} \hspace{0.01\textwidth}
\subfloat[{EVIL}]{\includegraphics[width=0.126\textwidth]{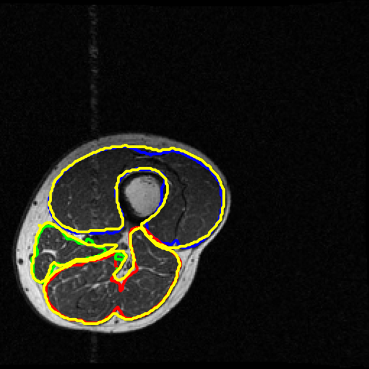}} \hspace{0.01\textwidth}
\subfloat[{VP-Net}]{\includegraphics[width=0.126\textwidth]{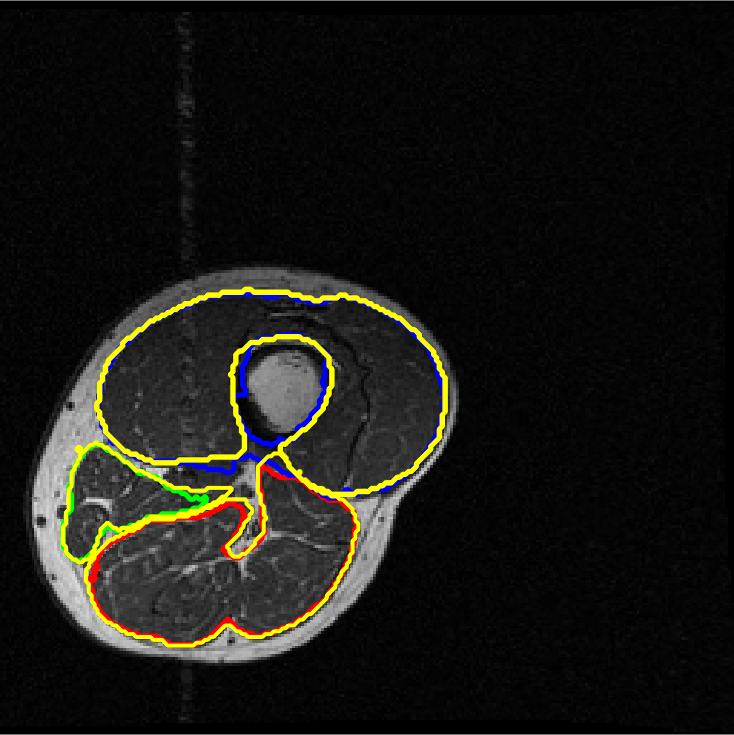}} 
\caption{Thigh muscle segmentation results obtained by different approaches. Predicted boundaries are indicated in red, blue, and green, and ground truth contours are highlighted in yellow.
}\label{thighimg}
\end{figure*}

Accurate segmentation of various thigh muscles plays a vital role in assessing musculoskeletal disorders like osteoarthritis.
However, this task is difficult because of unclear boundaries and the similar intensity and texture among the muscles.
The segmentation partitioned the thigh region into four main clusters: the quadriceps, hamstrings,  other muscle groups, and remaining tissues.
An example muscle MRI is presented in Fig. \ref{MR_example}.
Thigh muscle MR images contains 75 slices, each sized at $256 \times 256$. In our experiments, 45 labeled exams are used for training, with the remaining 30 reserved for validation. 
The training dataset includes $n$ fully annotated images, with pixel-wise labels unavailable for the remaining $45-n$.
Table \ref{thigh} provides the results. 
Fig. \ref{thighimg} illustrates the segmentation outputs produced by various models.
As shown in Fig. \ref{thighimg}, other methods struggle to accurately distinguish between different muscle categories. In contrast, the segmentation produced by our method aligns more closely with the ground truth and differentiates each muscle category more precisely, avoiding inter-class segmentation errors.

\begin{table}[htbp]
\centering
\caption{Quantitative results of different models on the thigh muscle MR images. Best results for VP-Net against other methods are shown in bold, while those for VP-Net\textsubscript{gt} are underlined.}\label{thigh}
\resizebox{1.0\textwidth}{!}{
\begin{tabular}{p{3.6cm}p{1.3cm}p{1.7cm}p{0.2cm}p{1.8cm}p{2.4cm}p{2.4cm}p{2.4cm}}
\hline
\multirow{2}{*}{Method} &\multicolumn{2}{l}{ \# of scans} &  &\multicolumn{4}{l}{Metrics} \\
\cline{2-3} \cline{5-8}
 & Labeled & Unlabeled & & Dice (\%)$\uparrow$ & Jaccard (\%)$\uparrow$ & ASD(voxel)$\downarrow$ & 95HD(voxel)$\downarrow$ \\
\hline
U-Net & 15 & 30 & & 85.92 & 77.01 & 6.40 & 21.51 \\
U-Net & 30 & 15 & & 92.57 & 86.55 & 1.83 & 6.41 \\
\hline
Self-training &  &  & & 87.45 & 78.92 & 4.98 & 17.27 \\
Curriculum &  &  & & 87.87 & 79.20 & 6.20 & 20.23 \\
UA-MT &  &  & & 89.09 & 81.63 & 5.17 & 20.15 \\
MC-Net+ & 15& 30 & & 87.63 & 79.52 & 8.46 & 25.36 \\
EVIL  & &  & &   89.59  &  82.03 &   4.18  &  17.03  \\
\textbf{VP-Net(Ours)} &  &  & &\textbf{91.44} & \textbf{84.97} & \textbf{3.17}& \textbf{11.20} \\
\textbf{VP-Net\textsubscript{gt}(Ours)} &&  & & \underline{92.67} &\underline{86.72} & \underline{2.59} & \underline{9.20} \\
\hline

Self-training &  &  & & 93.71 & 88.63 & 2.63 & 10.31 \\
Curriculum &  & & & 92.85 & 86.98 & 1.82 & 6.84 \\
UA-MT &  && & 92.99 & 87.77 & 4.18 & 12.98 \\
MC-Net+ & 30& 15   & & 93.01 & 87.44 & 8.24 & 22.16 \\
EVIL  & &  & &  93.44& 88.14 & 5.11 & 18.24\\
\textbf{VP-Net(Ours)} &  &  & &  \textbf{94.98} & \textbf{90.77} & \textbf{1.05} & \textbf{4.32} \\
\textbf{VP-Net\textsubscript{gt}(Ours)} &  &  & & \underline{95.09} & \underline{90.89} & \underline{0.95} & \underline{4.67} \\
\hline
\end{tabular}
}
\end{table}

\subsection{Results on PROMISE12}
\begin{figure*}[htbp]
    \centering
    \subfloat{\includegraphics[width=0.126\textwidth]{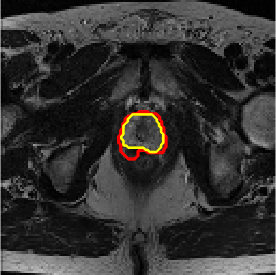}}\hspace{0.01\textwidth}
    \subfloat{\includegraphics[width=0.126\textwidth]{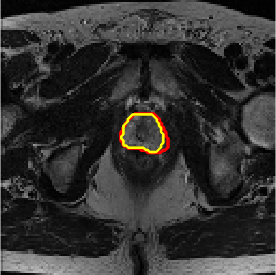}} \hspace{0.01\textwidth}
    \subfloat{\includegraphics[width=0.126\textwidth]{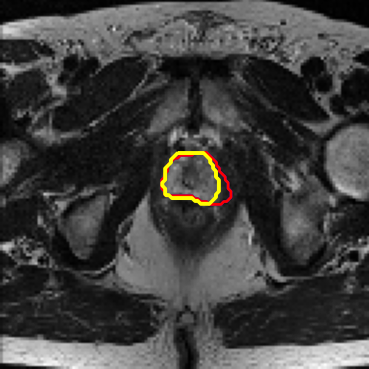}} \hspace{0.01\textwidth}
    \subfloat{\includegraphics[width=0.126\textwidth]{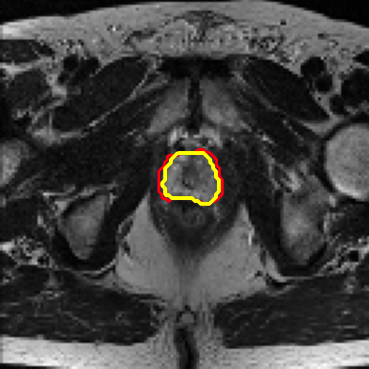}}\hspace{0.01\textwidth}
    \subfloat{\includegraphics[width=0.126\textwidth]{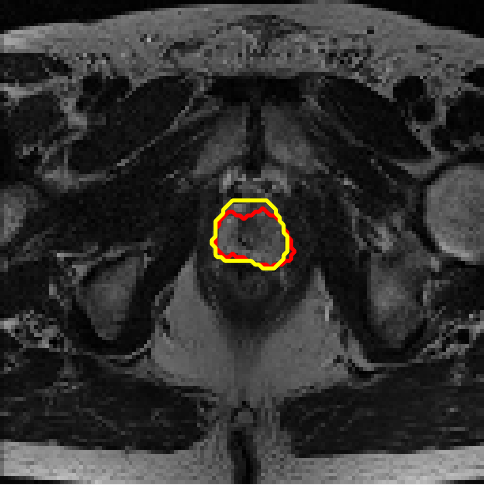}}\hspace{0.01\textwidth}
    \subfloat{\includegraphics[width=0.126\textwidth]{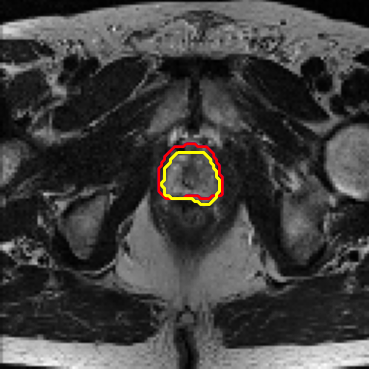}}\hspace{0.01\textwidth}
    \subfloat{\includegraphics[width=0.126\textwidth]{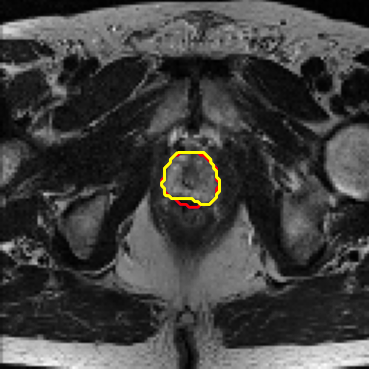}} \\
    \vspace{0.4pt}
    
     \subfloat{\includegraphics[width=0.126\textwidth]{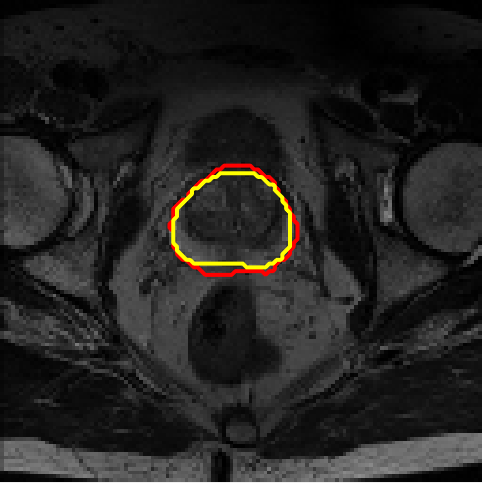}} \hspace{0.01\textwidth}
    \subfloat{\includegraphics[width=0.126\textwidth]{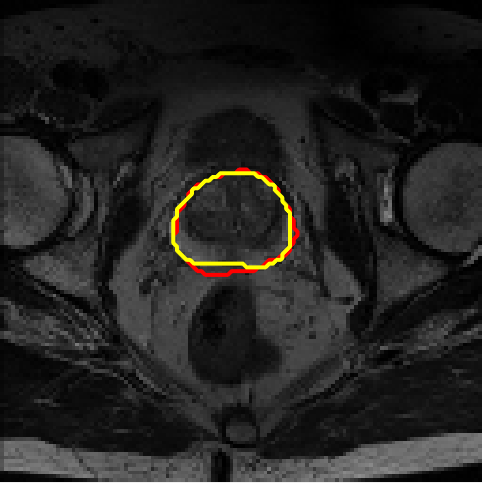}} \hspace{0.01\textwidth}
    \subfloat{\includegraphics[width=0.126\textwidth]{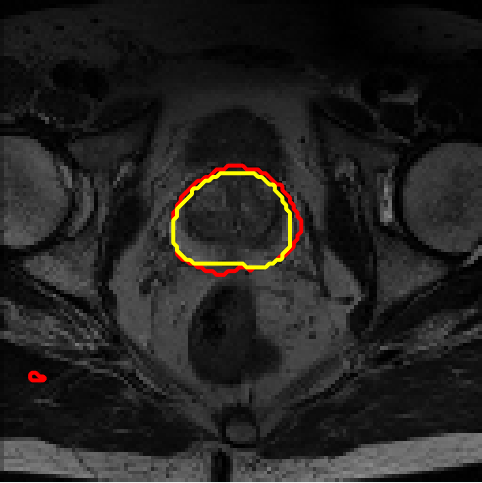}} \hspace{0.01\textwidth}
    \subfloat{\includegraphics[width=0.126\textwidth]{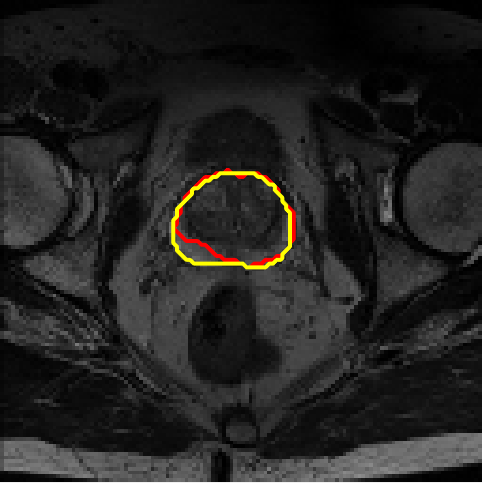}} \hspace{0.01\textwidth}
    \subfloat{\includegraphics[width=0.126\textwidth]{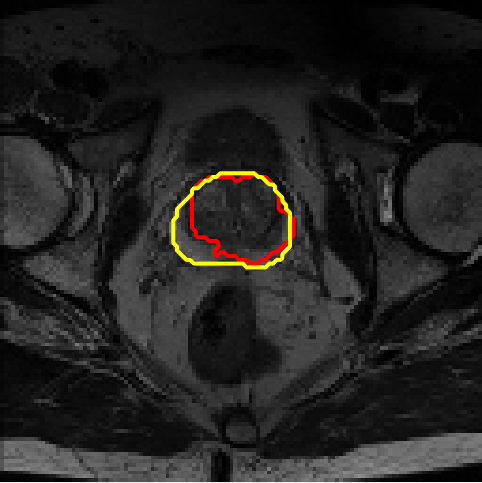}} \hspace{0.01\textwidth}
    \subfloat{\includegraphics[width=0.126\textwidth]{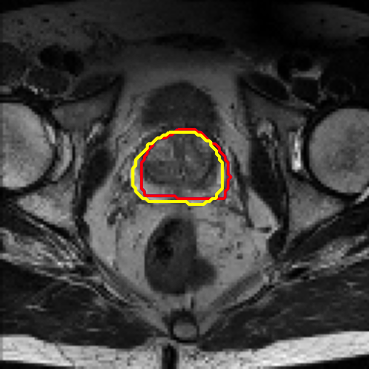}}\hspace{0.01\textwidth}
    \subfloat{\includegraphics[width=0.126\textwidth]{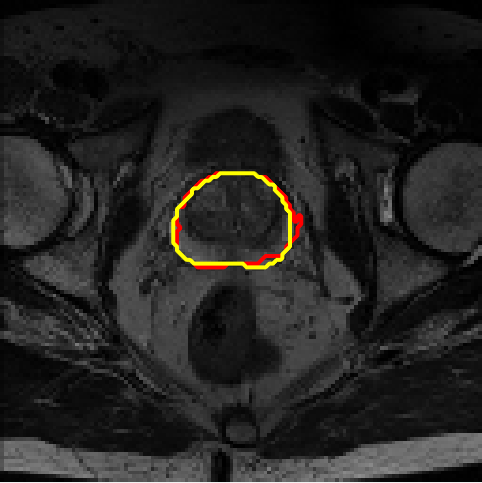}} \\
    
    \vspace{0.4pt}
    \setcounter{subfigure}{0}
     \subfloat[{U-Net}]{\includegraphics[width=0.126\textwidth]{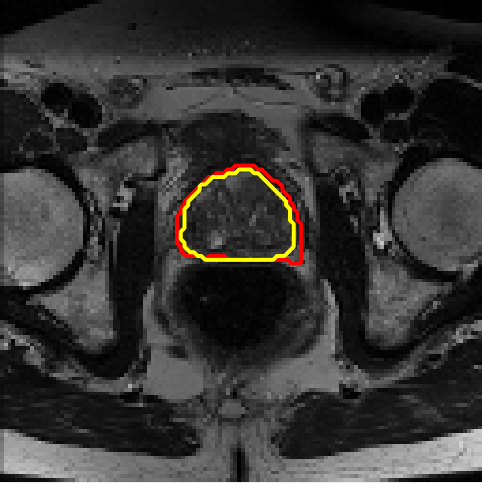}} \hspace{0.01\textwidth}
    \subfloat[{S-T}]{\includegraphics[width=0.126\textwidth]{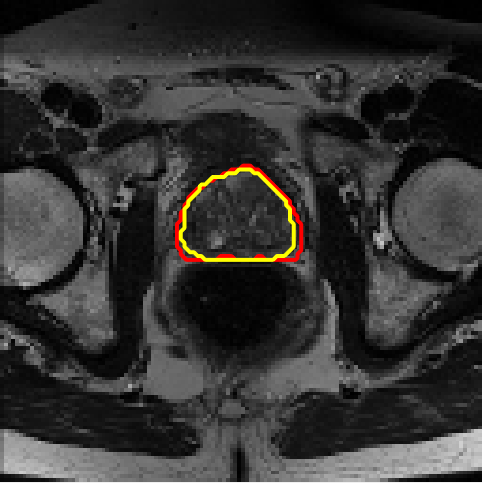}} \hspace{0.01\textwidth}
    \subfloat[{C-S}]{\includegraphics[width=0.126\textwidth]{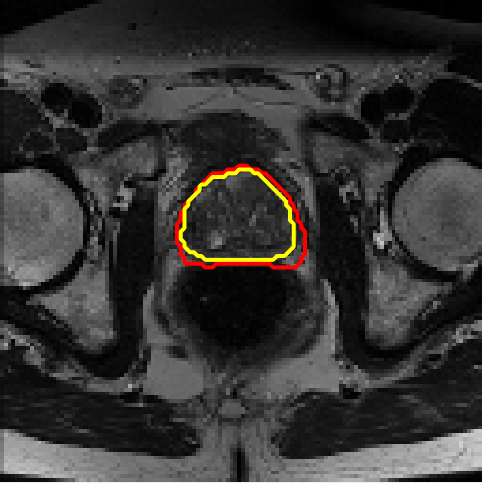}} \hspace{0.01\textwidth}
    \subfloat[{UA-MT}]{\includegraphics[width=0.126\textwidth]{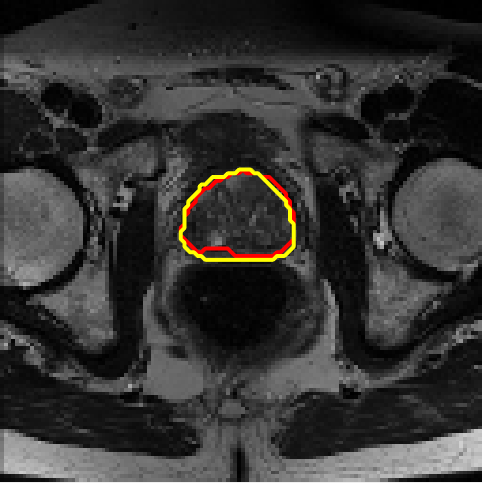}} \hspace{0.01\textwidth}    
    \subfloat[MC-Net+]{\includegraphics[width=0.126\textwidth]{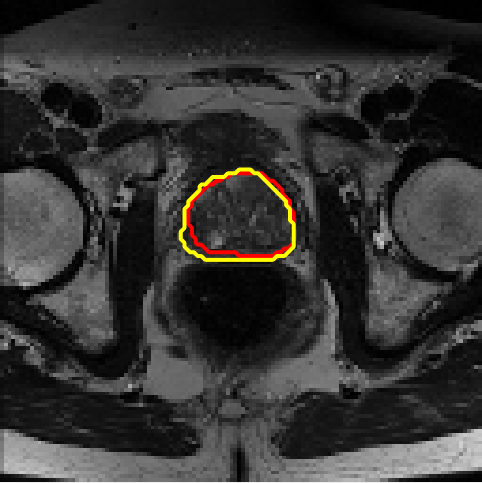}} \hspace{0.01\textwidth}
     \subfloat[EVIL]{\includegraphics[width=0.126\textwidth]{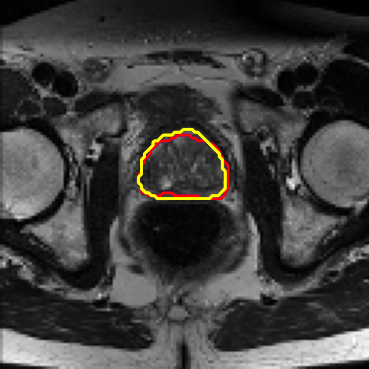}} \hspace{0.01\textwidth}
    \subfloat[{VP-Net}]{\includegraphics[width=0.126\textwidth]{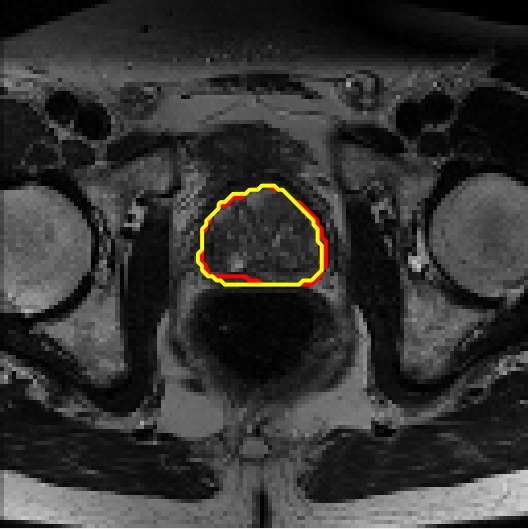}} \\
\caption{Example segmentation obtained by different methods on the PROMISE12 dataset. Red lines indicate predicted results, and yellow lines indicate ground truths.}\label{promiseimg}
\end{figure*}

We assess different semi-supervised methods on the PROMISE12 dataset \citep{promise}.
The PROMISE12 dataset contains 50 transverse T2-weighted MR images with the most anatomical detail. 35 and 15 cases are randomly selected for training and validation. The images were resized to $128\times 128$.
The training dataset includes $n$ fully annotated images, with pixel-wise labels unavailable for the remaining $35-n$.
Table \ref{promise} presents the results. Fig. \ref{promiseimg} offers a visual comparison of the segmentation results for different models and shows that the proposed method achieves more precise segmentation.

Evaluation on three different datasets indicates that the proposed method achieves the highest segmentation accuracy when the ground truth volumes of unlabeled images are provided to the VP-STD softmax layer (i.e., VP-Net\textsubscript{gt}), further indicating the importance of volume priors in the structure of semi-supervised segmentation networks.

\begin{table*}[htbp]
\centering
\caption{Quantitative results of different models on the PROMISE12 dataset. Best results for VP-Net against other methods are shown in bold, while those for VP-Net\textsubscript{gt} are underlined.}\label{promise}
\renewcommand{\arraystretch}{1.0}
\resizebox{1.0\textwidth}{!}{
\begin{tabular}{p{3.6cm}p{1.3cm}p{1.7cm}p{0.2cm}p{1.8cm}p{2.4cm}p{2.4cm}p{2.4cm}}
\hline
\multirow{2}{*}{Method} & \multicolumn{2}{c}{Scans used} & & \multicolumn{4}{c}{Metrics} \\
\cline{2-3} \cline{5-8}
 & Labeled & Unlabeled & & Dice (\%)$\uparrow$ & Jaccard (\%)$\uparrow$ & ASD(voxel)$\downarrow$& 95HD(voxel)$\downarrow$ \\
\hline
U-Net & 7 & 28 & & 84.54 & 77.56 & 6.77 & 19.28 \\
U-Net & 14 & 21 & & 87.98 & 81.75 & 2.62 & 8.12 \\
U-Net & 21 & 11 & & 91.18 & 86.03 & 1.36 & 3.95 \\
\hline
Self-training &  &  & & 88.40 & 81.78 & 2.70 & 8.34 \\
Curriculum &  &  & & 86.39 & 79.25 & 2.63 & 8.00 \\
UA-MT &  7&28  & & 88.06 & 81.71 & 1.90 & 5.35 \\
MC-Net+ &  &  & & 87.81 & 81.38 & 2.05 & 5.98 \\
EVIL & & & & 88.29     &81.80 &  2.90& 7.81\\
\textbf{VP-Net(Ours)} &  &  & & \textbf{88.76} & \textbf{82.14} & \textbf{1.68} & \textbf{5.06} \\
\textbf{VP-Net\textsubscript{gt}(Ours)} &  &  & & \underline{89.12} &\underline{83.18} &\underline{1.78} & \underline{4.57} \\
\hline
Self-training &  &  & & 89.87 & 84.04 & 1.34 & 4.09 \\
Curriculum &  &  & & 88.17 & 81.68 & 1.89 & 5.34 \\
UA-MT & 14 & 21 & & 88.37 & 82.51 & 1.79 & 5.04 \\
MC-Net+ &  &  & & 88.15 & 81.71 & 1.93 & 5.36 \\
EVIL & & & &  89.16   & 83.67&  1.87& 5.75\\
\textbf{VP-Net(Ours)} &  &  & & \textbf{90.35} & \textbf{84.43} & \textbf{1.36} & \textbf{4.13} \\
\textbf{VP-Net\textsubscript{gt}(Ours)} &  &  & & \underline{91.78} & \underline{87.11} & \underline{0.93} & \underline{3.07} \\
\hline
Self-training &  &  & & 91.61 & 86.59 & 1.22 & 3.62 \\
Curriculum &  &  & & 90.00 & 84.33 & 2.02 & 5.94 \\
UA-MT & 21 & 14  & & 91.40 & 86.29 & 1.19 & 3.65 \\
MC-Net+ &  &  & & 91.43 & 86.54 & 1.43 & 3.96 \\
EVIL & & & & 92.78    & 87.82& 1.32 & 4.06\\
\textbf{VP-Net(Ours)} &  &  & & \textbf{92.86} & \textbf{88.09} & \textbf{0.96} & \textbf{2.86} \\
\textbf{VP-Net\textsubscript{gt}(Ours)} & &  & & \underline{94.00} & \underline{89.83} & \underline{0.88} & \underline{2.75} \\
\hline
\end{tabular}
}
\end{table*}

\begin{table*}[htbp]
\centering
\caption{Performance comparison of \textbf{VP-Net} and \textbf{VP-Net}\textsubscript{rw}  across multiple datasets.}\label{ablation}
\renewcommand{\arraystretch}{1.0}
\resizebox{1.0\textwidth}{!}{
\begin{tabular}{p{3.4cm}p{2.0cm}p{1.3cm}p{1.5cm} p{0.3cm} p{1.8cm}p{2.2cm}p{2.2cm}p{2.2cm}}
\hline
{\multirow{2}{*}{Dataset} }& {\multirow{2}{*}{Methods} } &\multicolumn{2}{l}{Scans used}& & \multicolumn{4}{l}{Metrics} \\
  \cline{3-4}  \cline{6-9}
& & Labeled & Unlabeled  & &  Dice(\%)$\uparrow$  &  Jaccard(\%)$\uparrow$ & ASD(voxel)$\downarrow$ & 95HD(voxel)$\downarrow$ \\
\hline
\multirow{2}{*}{ACDC2017}
& \multirow{1}{*}{VP-Net\textsubscript{rw}} 
&  &  & &77.09 & 69.46& 4.41& 7.28\\
& \multirow{1}{*}{VP-Net\textsubscript{emp}} 
& 60& 80& & 75.94 & 68.52 & 3.98 & 6.65 \\
& \multirow{1}{*}{VP-Net} & 
 &  & &77.28 & \textbf{69.69}& \textbf{3.57} &\textbf{5.82} \\

\hline
\multirow{2}{*}{thigh muscle images} 
& \multirow{1}{*}{VP-Net\textsubscript{rw}} 
&  &  & & 91.08 & 84.01 & 3.77&15.12 \\
& \multirow{1}{*}{VP-Net\textsubscript{emp}} 
& 15& 30&   & 86.99 &  78.40 &  8.16 & 28.14 \\

& \multirow{1}{*}{VP-Net} & 
 &  & &\textbf{91.44} &\textbf{84.97}& \textbf{3.17} & \textbf{11.20}\\
\hline
\multirow{2}{*}{PROMISE12}
& \multirow{1}{*}{VP-Net\textsubscript{rw}} 
&  & & & 88.07 & 81.64&1.87 &5.29 \\
& \multirow{1}{*}{VP-Net\textsubscript{emp}} 
& 7 & 28& &  85.67 &  78.20& 4.23 &9.40\\
& \multirow{1}{*}{VP-Net} & 
 &  &
&\textbf{88.76} & \textbf{82.14}& \textbf{1.68} & \textbf{5.06}\\
\hline
\end{tabular}}

\end{table*}

\subsection{Ablation experiment}
To evaluate the contribution of the weak volume distribution prior, we perform an ablation study by removing the loss term $\mathcal{L}_{W}$ from our semi-supervised segmentation framework.
When this prior is removed (denoted as VP-Net\textsubscript{rw}), a performance drop is observed across all datasets (ACDC2017, thigh muscle MR images, and PROMISE12), as shown in Table \ref{ablation}.
These results indicate that incorporating the proposed weak prior improves segmentation performance within the semi-supervised framework.

To validate the effectiveness of the proposed learnable volume prediction mechanism, we conduct ablation experiments by replacing our volume prediction with the empirical average volume $\mathbf{V}_{emp}$ used in \citep{liuj2022a}, which we refer to as VP-Net\textsubscript{emp}. Specifically, we disable the regression network and directly fed  $\mathbf{V}_{emp}$ (pre-calculated from the ground truth labels of the training set) to the VP-STD softmax layer during both training and testing phases, while keeping all other network components unchanged. As demonstrated in Table \ref{ablation}, our method with adaptive volume prediction consistently surpasses the $\mathbf{V}_{emp}$-based approximation across all test datasets. These results confirm that our learnable volume prediction mechanism achieves more accurate volume estimation compared to fixed empirical averages from the training set.

\section{Conclusion and discussion}
\label{conclusion}
\noindent 

In this paper, we propose a semi-supervised medical image segmentation framework that incorporates TD spatial regularization and volume priors. The framework integrates a strong image-scale volume prior into the backbone segmentation network, while a regression network predicts target region volumes for unlabeled images, effectively regularizing the segmentation process. The loss function for unlabeled samples is constructed using weak dataset-scale priors based on the similarity of volume distributions. 
Experimental results demonstrate that our proposed method achieves more accurate segmentation compared to some SOTA semi-supervised segmentation networks. 

By integrating spatial regularization and volume priors, our framework enhances the mathematical interpretability of segmentation process, thereby facilitating more reliable analyses of tissues and organs. Furthermore, our method provides opportunities to incorporate other spatial regularization priors in semi-supervised segmentation in the future.
Although the backbone of our framework is currently based on U-Net, it is flexible and can be replaced with more advanced models, such as the Segment Anything Model (SAM) \citep{segmentany}. This flexibility enables potential improvements in segmentation performance.

\section{Acknowledgments}
This work was supported by the National Natural Science Foundation of China (No. 12371527), the Beijing Natural Science Foundation (No. 1232011), and the Fundamental Research Program of Shanxi Province (No. 202403021222002).

\bibliographystyle{elsarticle-harv}
\bibliography{references}


\end{document}